\documentclass[pdflatex,sn-mathphys-num,iicol]{sn-jnl}%

\usepackage{graphicx}%
\usepackage{multirow}%
\usepackage{amsmath,amssymb,amsfonts}%
\usepackage{amsthm}%
\usepackage{mathrsfs}%
\usepackage[title]{appendix}%
\usepackage[dvipsnames,table]{xcolor}%
\usepackage{textcomp}%
\usepackage{manyfoot}%
\usepackage{booktabs}%
\usepackage{algorithm}%
\usepackage{algorithmicx}%
\usepackage{algpseudocode}%
\usepackage{listings}%
\usepackage{calc}

\usepackage{lipsum}
\usepackage{tikz}
\usetikzlibrary{arrows.meta,positioning,shapes.geometric,calc}
\usepackage{pgfplots}
\usepackage{pgffor}
\usepackage{tabularx}
\usepgfplotslibrary{groupplots,fillbetween}
\pgfplotsset{compat=1.18}
\usepackage{xparse}

\usepackage{orcidlink}

\usepackage{enumitem}
\usepackage{fancyhdr}
\usepackage{balance}

\usepackage{pifont}%
\usepackage{colortbl}
\usepackage{caption}
\usepackage{subcaption}
\usepackage{tablefootnote}
\usepackage{relsize}
\usepackage[accsupp]{axessibility} %
\usepackage{etoolbox}

\usepackage{eso-pic}

\usepackage[normalem]{ulem}
\usepackage[leftcaption]{sidecap} %

\usepackage[capitalize]{cleveref}
\crefname{section}{Sec.}{Secs.}
\Crefname{section}{Section}{Sections}
\Crefname{table}{Table}{Tables}
\crefname{table}{Tab.}{Tabs.}
\crefname{algorithm}{Alg.}{Algs.}
\Crefname{algorithm}{Algorithm}{Algorithms}

\theoremstyle{thmstyleone}%

\theoremstyle{thmstyletwo}%

\theoremstyle{thmstylethree}%

\newcommand{\ProjectWebsite}[0]{\href{https://MiraPurkrabek.github.io/BBox-Mask-Pose/}{project website}}
\newcommand{\IoUMax}[0]{\textit{IoU}_{\textit{max}}}

\newcommand{\cmark}{\ding{51}}%
\newcommand{\xmark}{\ding{55}}%

\definecolor{LightCyan}{rgb}{0.7,1,1}
\definecolor{LightYellow}{rgb}{1,1,0.7}
\definecolor{LighterYellow}{rgb}{1,1,0.9}
\definecolor{LightOrange}{rgb}{1,0.8,0.5}
\definecolor{DarkGreen}{rgb}{0.0,0.5,0.0}
\definecolor{Burgundy}{rgb}{0.5, 0, 0.125}
\definecolor{HunterGreen}{rgb}{0.207, 0.367, 0.23}
\definecolor{PastelGreen}{rgb}{0.9, 1.0, 0.9} %
\definecolor{LightBlue}{RGB}{220, 245, 255}
\definecolor{DarkGray}{RGB}{120, 120, 120}
\definecolor{BckPlayerBlue}{RGB}{0, 180, 255}
\definecolor{ForePlayerMagenta}{RGB}{255, 0, 255}
\definecolor{BrightBlue}{RGB}{0, 150, 255}
\definecolor{CobaltBlue}{RGB}{0, 71, 17}
\definecolor{BlueGreen}{RGB}{8, 143, 143}

\definecolor{Magenta}{RGB}{252, 2, 207}
\definecolor{Cyan}{RGB}{0, 127, 255}
\definecolor{Orange}{RGB}{246, 161, 5}
\definecolor{DarkOrange}{RGB}{230,126,34}

\definecolor{bmpColorMagenta}{RGB}{255, 0, 255}
\definecolor{bmpColorBlue}{RGB}{0, 128, 255}
\definecolor{bmpColorOrange}{RGB}{255, 128, 0}
\definecolor{bmpColorGreen}{RGB}{128, 191, 128}
\definecolor{bmpColorRed}{RGB}{182, 1, 65}

\makeatletter
\patchcmd{\@maketitle}%
  {{\printkeywords\par}}%
  {{\printkeywords\par}%
   \begin{center}
     \begin{minipage}{\dimexpr\textwidth-48pt\relax}
       \centering
        \includegraphics[height=2.7cm]{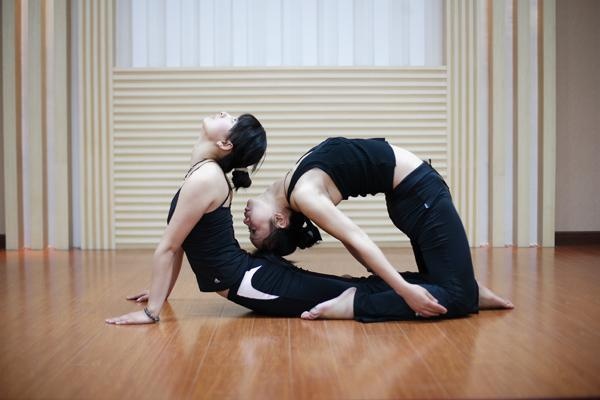}
        \hfill
        \includegraphics[height=2.7cm]{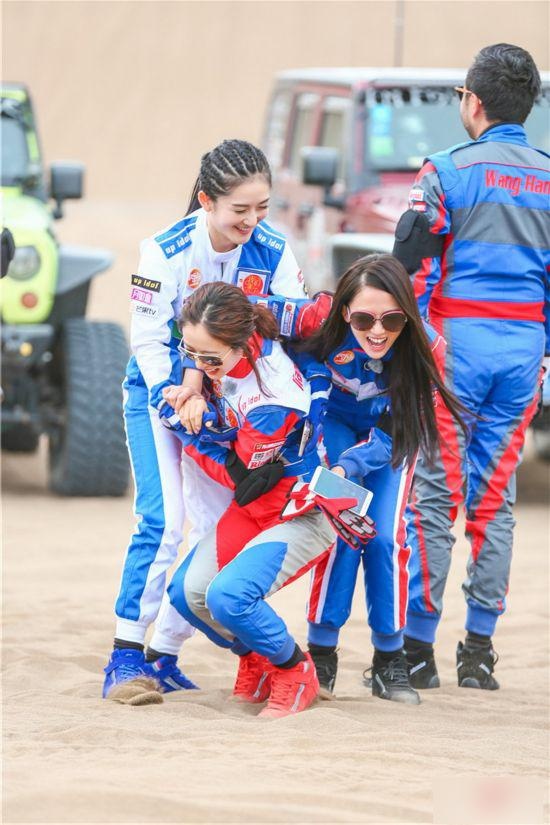}
        \hfill
        \includegraphics[height=2.7cm]{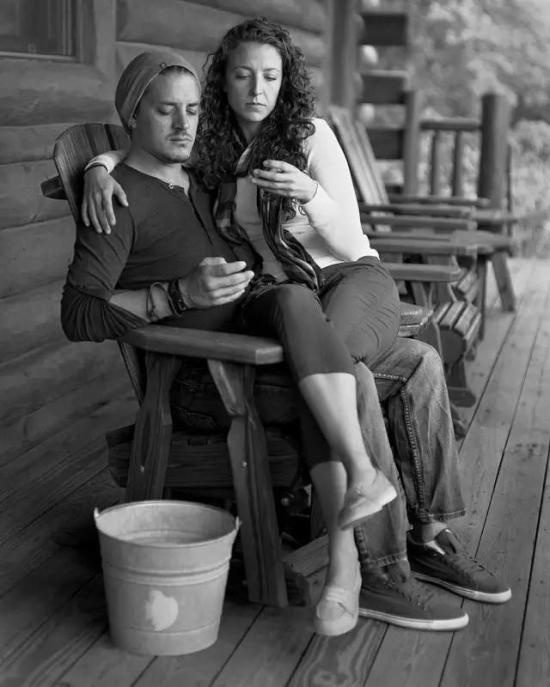}
        \hfill
        \includegraphics[height=2.7cm]{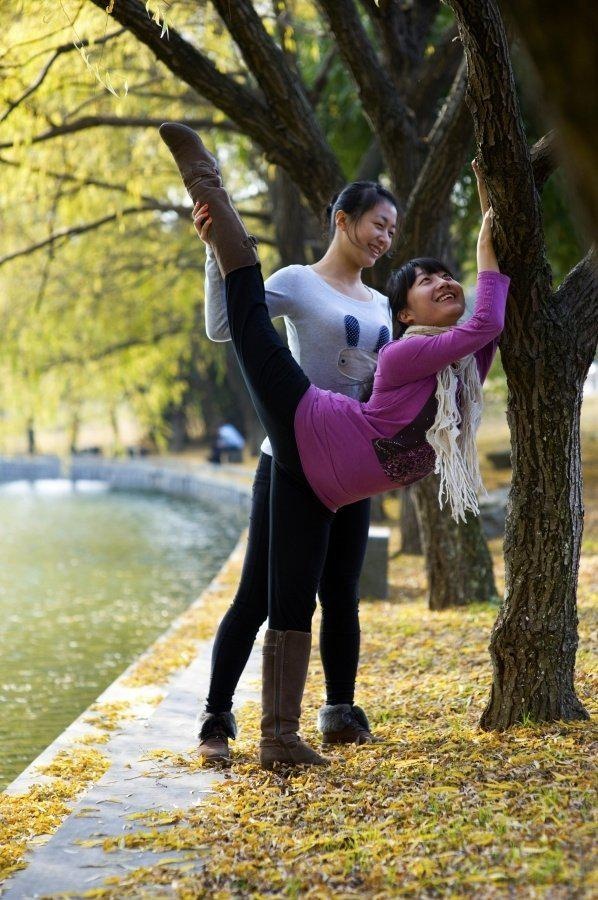}
        \hfill
        \includegraphics[height=2.7cm]{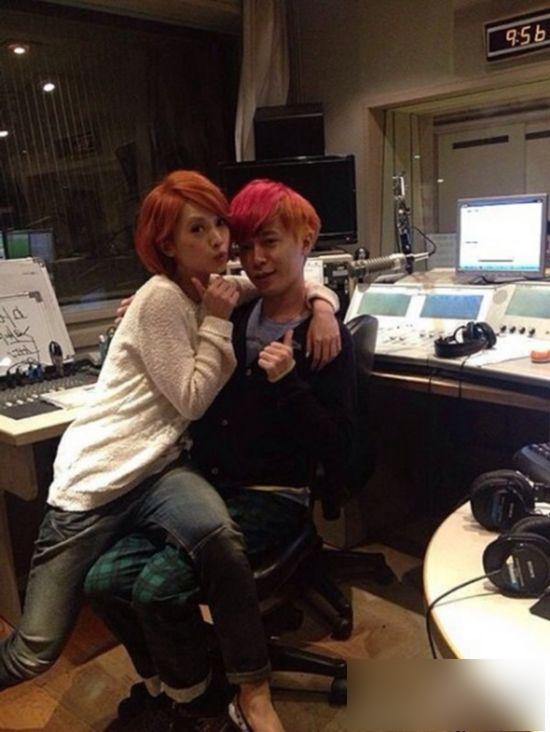}
        \hfill
        \includegraphics[height=2.7cm]{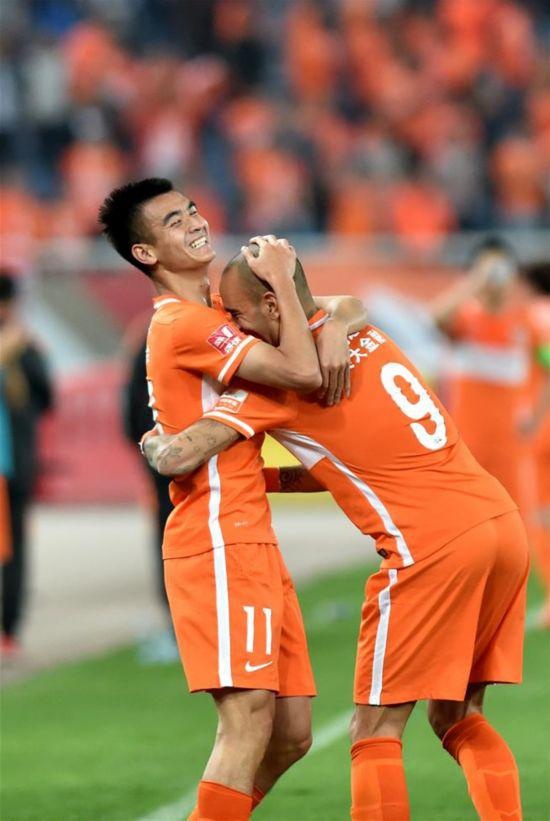}
        
        \includegraphics[height=2.7cm]{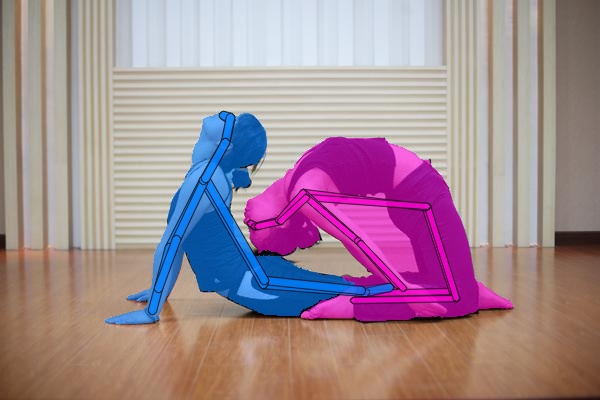}
        \hfill
        \includegraphics[height=2.7cm]{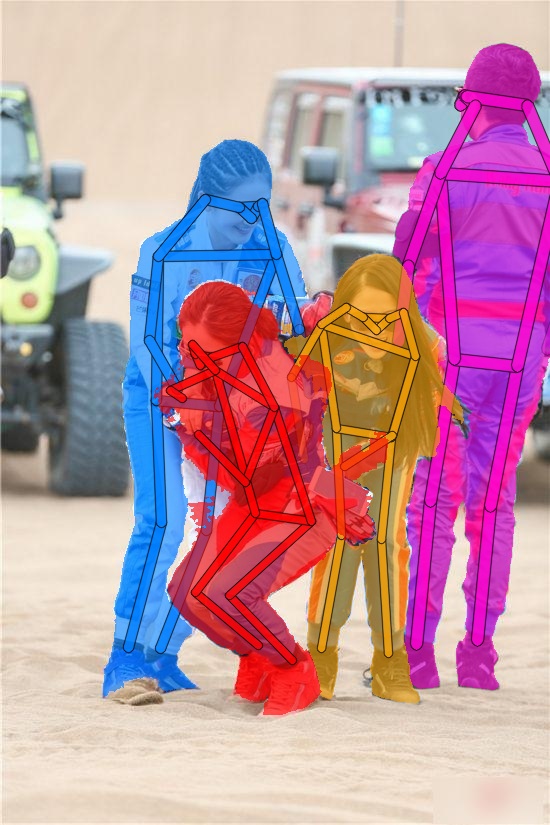}
        \hfill
        \includegraphics[height=2.7cm]{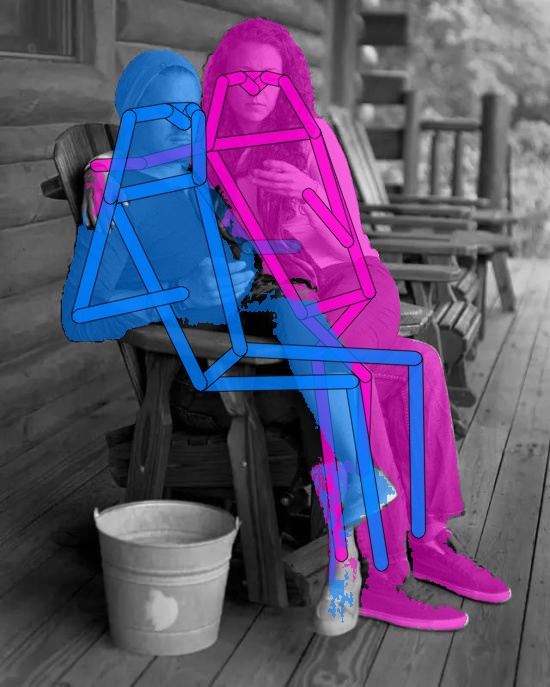}
        \hfill
        \includegraphics[height=2.7cm]{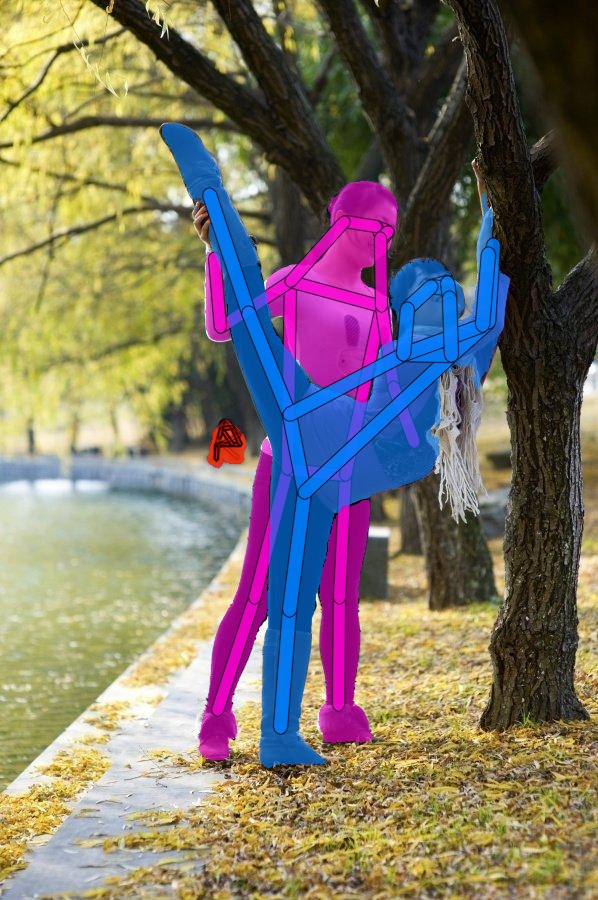}
        \hfill
        \includegraphics[height=2.7cm]{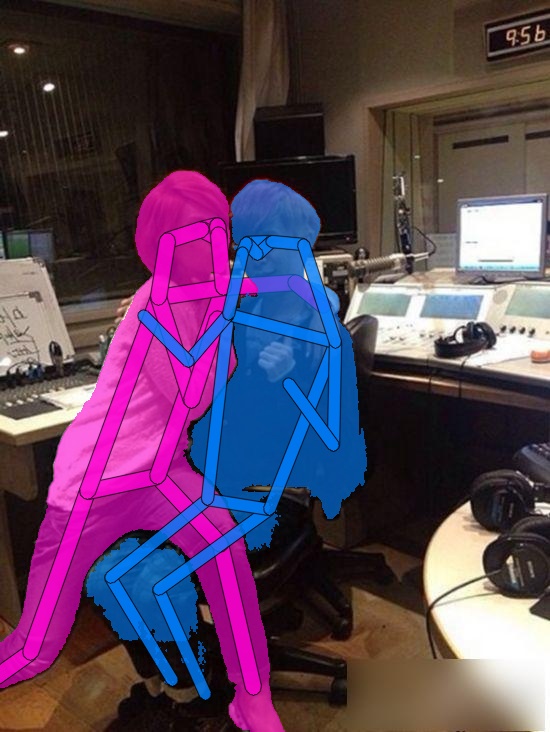}
        \hfill
        \includegraphics[height=2.7cm]{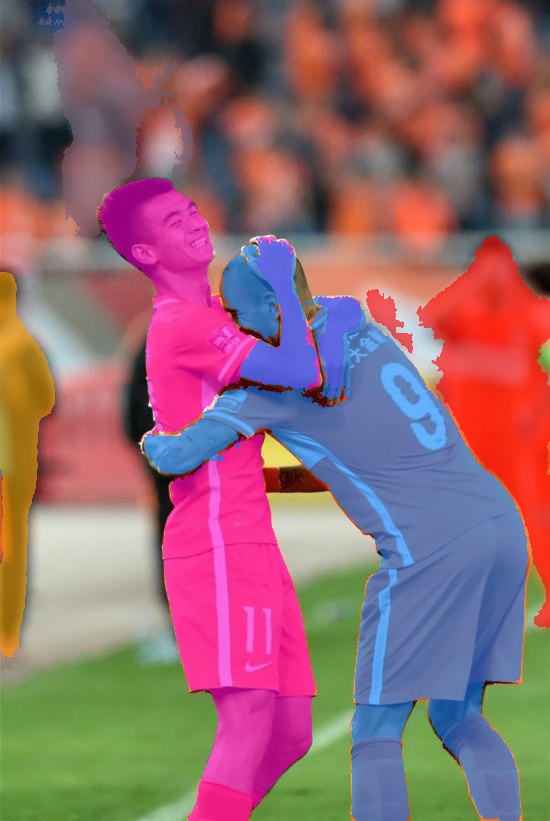}
        
        \includegraphics[height=2.7cm]{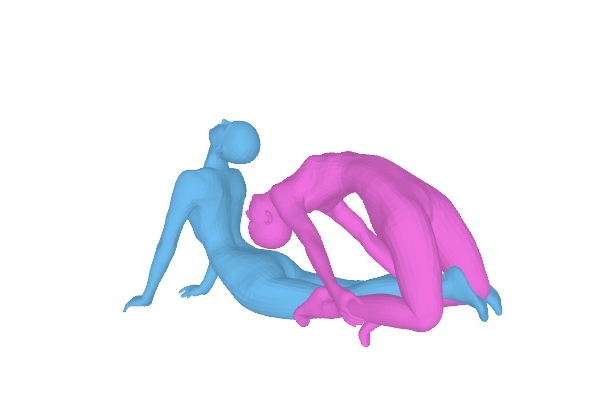}
        \hfill
        \includegraphics[height=2.7cm]{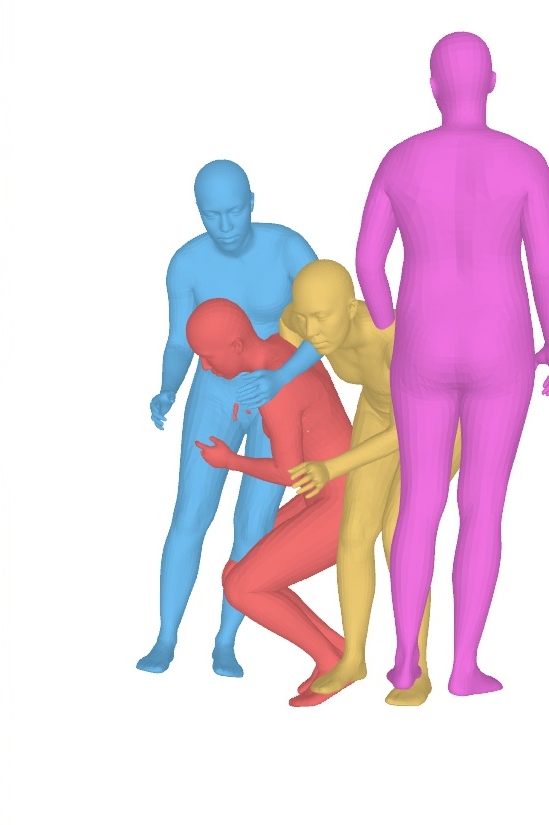}
        \hfill
        \includegraphics[height=2.7cm]{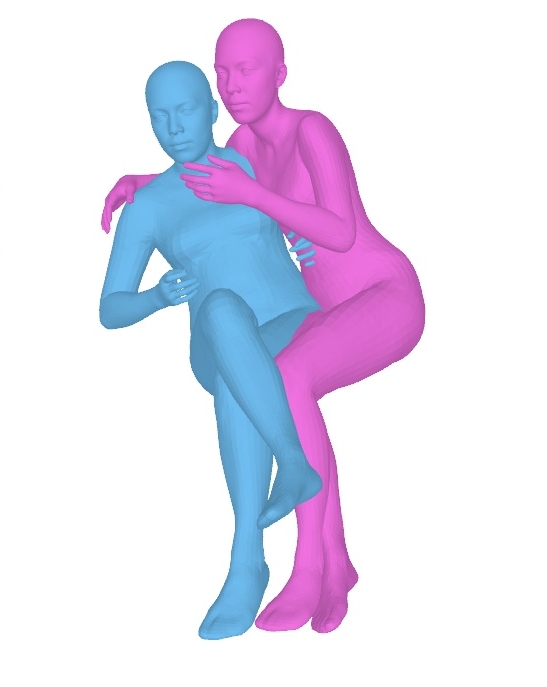}
        \hfill
        \includegraphics[height=2.7cm]{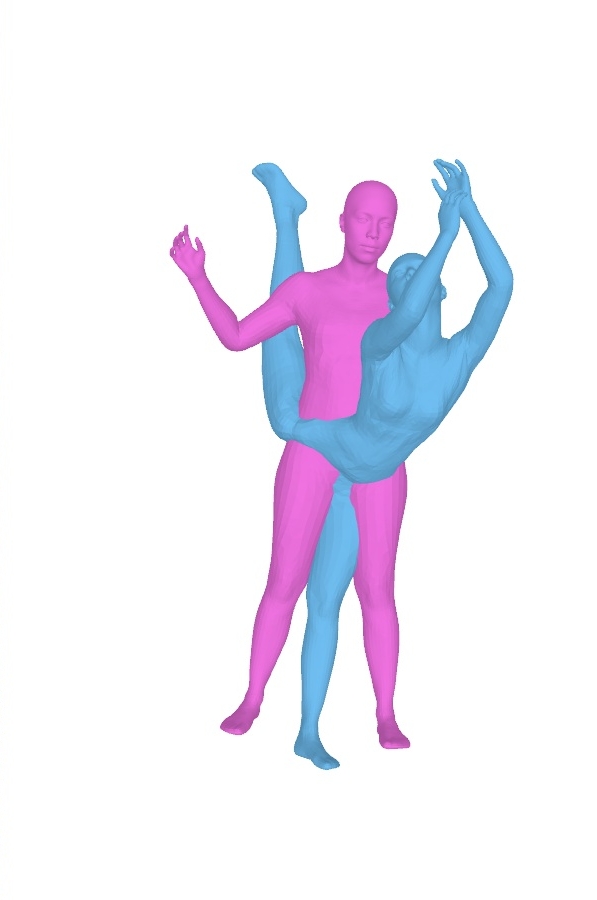}
        \hfill
        \includegraphics[height=2.7cm]{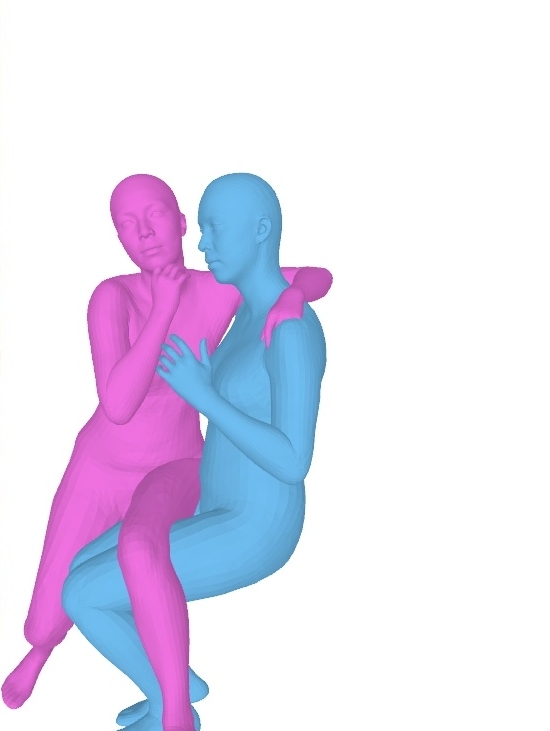}
        \hfill
        \includegraphics[height=2.7cm]{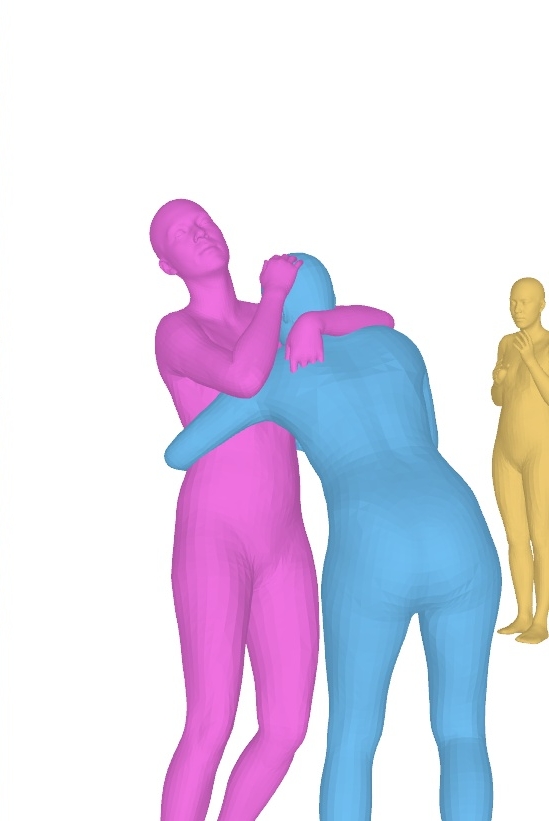}
       \captionof{figure}{%
         \textbf{BBoxMaskPose v2} establishes a new SOTA detection, segmentation, and 2D pose estimation in crowded scenes (row 2).
         Combined with SAM-3D-Body \cite{SAM-3D-Body}, it enables 3D human reconstruction even in scenes with complex interactions.
       }
       \label{fig:teaser}
     \end{minipage}
   \end{center}
  }%
  {}{}%
\makeatother

\begin{document}

\AddToShipoutPictureFG*{%
  \AtPageUpperLeft{%
    \raisebox{-2.0cm}{%
      \makebox[\paperwidth]{%
        \begin{minipage}{0.92\paperwidth}
          \centering
          \color{gray}

          {\fontsize{7}{8}\selectfont

          This preprint has not undergone peer review or any post-submission improvements or corrections.\\
          The Version of Record of this article is published in the International Journal of Computer Vision (IJCV) and is available online at \texttt{http://doi.org/10.1007/s11263-026-03113-6}.\\
          \vspace{1em}
          The full accepted manuscript, including additional experiments and ablations, is freely available on the \ProjectWebsite.

          \par}

        \end{minipage}%
      }%
    }%
  }%
}

\title[Article Title]{
BBoxMaskPose v2: Expanding Mutual Conditioning to 3D
}

\author*[1]{\fnm{Miroslav} \sur{Purkrabek} \orcidlink{0009-0000-6142-6492}*}\email{purkrmir@fel.cvut.cz}

\author[]{\fnm{Constantin} \sur{Kolomiiets}}\email{\nomail}

\author[]{\fnm{Jiri} \sur{Matas} \orcidlink{0000-0003-0863-4844}}\email{\nomail}

\affil[]{
\centering
\orgdiv{Visual Recognition Group},
\orgname{Czech Technical University in Prague,\\
}
\vspace{-2.8em}
}

\abstract{
Most 2D human pose estimation benchmarks are nearly saturated, with the exception of crowded scenes.
We introduce PMPose, a top-down 2D pose estimator that incorporates the probabilistic formulation and the mask-conditioning.
PMPose improves crowded pose estimation without sacrificing performance on standard scenes. 
Building on this, we present BBoxMaskPose v2 (BMPv2) integrating PMPose and an enhanced SAM-based mask refinement module.
BMPv2 surpasses state-of-the-art by 1.5 average precision (AP) points on COCO and 6 AP points on OCHuman, becoming the first method to exceed 50 AP on OCHuman.
We demonstrate that BMP’s 2D prompting of 3D model improves 3D pose estimation in crowded scenes and that advances in 2D pose quality directly benefit 3D estimation. %
Results on the new OCHuman-Pose dataset show that multi-person performance is more affected by pose prediction accuracy than by detection.\\
The code, models, and data are available on
the \ProjectWebsite.
\vspace{-0.5em}
}
\keywords{Human Pose Estimation, Person Detection, Instance Segmentation, Mutual Conditioning\vspace{-0.5em}}

\maketitle

\section{Introduction}
\label{sec:intro}

Human pose estimation (HPE) research is gradually shifting from 2D to 3D.
However, state-of-the-art 3D pose estimators~\cite{PromptHMR, SAM-3D-Body} rely on accurate 2D input, whether from detections, masks, or points.
Thus, improving the 2D HPE remains essential for reliable 3D human body modelling.

Multi-person interaction, where individuals are so close that their bounding boxes are almost the same, is one of the main challenges in 2D and 3D.
Real-world data for such cases are limited~\cite{CrowdHuman,CrowdPose,CIHP,pose2seg}.
Existing ``crowded'' datasets vary in difficulty~--~datasets such as~\cite{CrowdHuman, CrowdPose, CIHP} include individuals close to each other, but lack the interactions present in OCHuman~\cite{pose2seg}.

BMPv1~\cite{BMPv1} (ICCV 2025) addresses joint detection and pose estimation in multi-person scenes and achieves strong results in crowded scenes.
Its core is a self-correcting loop of three lightweight models -- a detector, a pose estimator, and a segmentation model -- conditioned on each other to iteratively refine their predictions.
BMPv1 shows that mutual conditioning between different human-body representations is effective in crowded scenes.
However, the prompting of SAM~\cite{SAM2} within BMP is complex, often leads to over-segmentation, and makes the mask refinement stage the dominant failure mode.

ProbPose~\cite{ProbPose} (CVPR 2025) addresses different limitations of top-down human pose estimation through probabilistic keypoint modeling.
It predicts visibility, presence probability, and expected OKS\footnote{Object Keypoint Similarity \cite{COCO}} for each keypoint, and uses crop augmentation to improve robustness near image borders.
These outputs are attractive for BMP because they are useful not only for pose localization, but also for selecting informative prompts for mask refinement.

The work on BMPv1 \cite{BMPv1} and ProbPose \cite{ProbPose} is complementary.
BMPv1 use mask conditioning, which is important for disentangling individuals in crowds, but it lacks rich probabilistic outputs for calibration and prompt selection.
ProbPose provides such outputs, but it is not designed for explicit instance-mask conditioning inside the crowded BMP loop.
Crowded mutual conditioning requires both properties at once.

This paper introduces \textbf{BMPv2}, an upgrade of BMPv1 that consistently improves pose estimation and segmentation performance and builds on BMPv1~\cite{BMPv1} (ICCV 2025), ProbPose~\cite{ProbPose} (CVPR 2025), and SAM-pose2seg~\cite{SAM-pose2seg-report} (CVWW 2026).
Its new contributions are PMPose, the BMPv2 system integration, the higher-precision BMPv2+ variant, OCHuman-Pose, and broader analysis of prompting, runtime, and crowded-scene data.

BMPv2 extends BMPv1 with two new components: PMPose and SAM-pose2seg.
By incorporating these changes, BMPv2 sets a new SOTA performance on both standard scenes (i.e. on COCO~\cite{COCO}) and crowds (i.e. on OCHuman~\cite{pose2seg}) for 2D pose estimation and instance segmentation.
The improved mask refinement from SAM-pose2seg further enables an additional PMPose refinement, resulting in \textbf{BMPv2+}, which targets high-precision scenarios at the cost of increased computation.

\textbf{PMPose} is a new top-down pose estimation model introduced as part of BMPv2.
It combines mask conditioning with probabilistic keypoint modeling in a single architecture.
It is suitable both for the BMP loop and as a standalone 2D top-down human pose estimator.
PMPose outperforms previous non-iterative pose estimation methods in both standard and crowded scenes.

BMPv2 uses \textbf{SAM-pose2seg}, a version of SAM~\cite{SAM2} adapted for human pose prompts, for mask refinement.
It reduces over-segmentation and replaces the complex prompting strategy of BMPv1.
It also makes direct use of the visibility-related information provided by PMPose.

We also analyze OCHuman and CIHP, two crowded-scene benchmarks with different weaknesses.
OCHuman contains close interactions but incomplete annotations, while CIHP provides precise masks but no pose annotations.
To address the main limitation of OCHuman, we introduce \textbf{OCHuman-Pose}, which adds COCO-style keypoint annotations to 2739 previously bounding-box-only instances in the original images.
Results on OCHuman-Pose suggest that detection is less dominant than originally implied by OCHuman and that pose estimation remains a major bottleneck.

Finally, we use BMPv2 to prompt SAM-3D-Body~\cite{SAM-3D-Body} and analyze 3D estimation in challenging crowded scenes.
This study shows that stronger 2D outputs improve 3D human body estimation in crowds.
This supports the main motivation of the paper: robust 2D understanding remains a necessary part of in-the-wild 3D pipelines.

The main contributions of this paper are:
\begin{enumerate}
    \item \textbf{PMPose}: a family of mask-conditioned probabilistic 2D top-down pose estimators that improve both crowded-scene and standard-scene performance (\cref{sec:method-PMPose}).
    
    \item \textbf{BMPv2} and \textbf{BMPv2+}: improved crowded-scene methods integrating PMPose and SAM-pose2seg that set a new SOTA for pose estimation and instance segmentation (\cref{sec:method,sec:bmp-plus,sec:exp}).
    
    \item \textbf{OCHuman-Pose dataset}: an extension of OCHuman~\cite{pose2seg} with additional keypoint annotations for previously missing instances (\cref{sec:data-OCH}).
    
    \item \textbf{Analysis of BMP components}: ablations of PMPose, SAM-pose2seg prompting, runtime, and crowded-scene data weaknesses (\cref{sec:exp-abl-PMPose,sec:exp-abl-runtime,sec:data}).

    \item \textbf{Downstream 3D prompting study}: analysis of how improved 2D outputs affect 3D human body estimation in crowds (\cref{sec:exp-3D}).
    
\end{enumerate}

\section{Related Work}
\label{sec:related}

\noindent
\textbf{Detection-Based Segmentation:}
The most direct approach to human instance segmentation uses a detector~\cite{coDETR, YOLOX, YOLOv3, RTMDet, ViTDet, HRNet, ConvNeXt, HRNet}.
A detector takes an image as input and outputs instance masks and class labels.
Detectors work well in scenes similar to the training data, but struggle when instances overlap heavily.
Under severe occlusion, they often merge multiple people into one mask, or assign different body parts to different instances.
In crowded scenes, standard non-maximum suppression (NMS) can remove true positives, leading to missed instances. Relaxing NMS (e.g., higher IoU thresholds) can recover these instances, but typically increases false positives in non-crowded scenes.
\cite{PoseNMS} proposed to suppress detection not on bounding box level but after pose estimation, which results in longer inference times since false positive detections have to first be processed by the pose estimator before being suppressed.

\vspace{1em}
\noindent
\textbf{Semantics-Aware Prompted Segmentation:}
Several works~\cite{GroundedSAM, SAMWise, SA-SAM, OpenWorldSAM} inject semantics into SAM~\cite{SAM2}, similarly to our SAM-pose2seg.
They take an image and a text prompt as input and detect and segment instances.
This setting is often referred to as \textit{open-vocabulary} or \textit{zero-shot} segmentation.
While these models capture semantics and segment entire instances, they do not provide localized cues and offer no control over which instance is chosen and, therefore, are unsuitable for iterative loops such as BMP~\cite{BMPv1}.

\vspace{1em}
\noindent
\textbf{Pose-Guided Instance Segmentation}
is the most specific form of prompted segmentation.
The model takes an image and a detected or ground-truth human pose and segments the corresponding person.
These methods specialize in humans, sacrificing the generality of open-vocabulary or generic segmenters for robustness and precise, localized control.
Test-time optimization methods~\cite{ttattg, PoseSegTTA} refine predictions at inference without training.
While training-free, they are computationally expensive and slow at test time.
Standard pose-guided human segmentation methods~\cite{pose2body, pose2instance, pose2seg, PosePlusSeg, PoSeg, MultiPoseSeg, ParsingRCNN} rely on small training datasets and thus generalize poorly, especially in crowded scenes.
The closest work related to SAM-pose2seg is CrowdSAM~\cite{CrowdSAM}, which builds a framework around SAM to automatically annotate bounding boxes in crowds.
CrowdSAM uses SAM and DINO~\cite{DINOv2} as external tools and optimizes prompts, similar to~\cite{BMPv1}.
In contrast, our SAM-pose2seg is end-to-end and, when used in an iterative loop, outperforms CrowdSAM, as shown already in~\cite{BMPv1}.

\vspace{1em}
\noindent
\textbf{2D Human Pose Estimation:}
There are two main approaches: top-down and detector-free.
Detector-free can be further divided into single-stage~\cite{POET,PETR,AdaptivePose++,CID}, bottom-up~\cite{DEKR,OpenPose,BottomUpSeg} and hybrid~\cite{BUCTD}.

Top-down methods~\cite{ViTPose,HRNet,SWIN,HRFormer,MIPNet,PVTv2} use a person detector to detect bounding boxes and estimate one skeleton for each bounding box.
They leverage big progress in human detection and specialize in understanding of human structure.
Top-down methods are the most successful on datasets such as COCO, MPII or AIC but struggle on crowded datasets (e.g. OCHuman) due to low-quality detections.
Most notably, ViTPose~\cite{ViTPose} combines multiple datasets into one strong backbone and sets a strong baseline, setting up state-of-the-art performance on most datasets.
While conditioning pose estimation on bounding boxes (bbox-to-pose; standard top-down approach) is well researched, conditioning pose on masks (mask-to-pose) was explored only through MaskPose~\cite{BMPv1}.

While the vast majority of top-down approaches are heatmap-based, some models~\cite{DeepPose, PoseRegression, RLE} predict keypoints through regression or regression-focused heatmaps~\cite{IntegralPose, IntegralPose2}.
Important for this work is ProbPose~\cite{ProbPose} -- a heatmap-based method modelling the underlying probability distribution.
ProbPose is the first model to estimate probabilities of keypoints being out of the image along with their visibilities.
As will be shown in this work, such estimates are useful for pose-guided instance segmentation in the BMP loop.

Detector-free models do not achieve SOTA performance on COCO but are superior to top-down methods on OCHuman as they are specialized in decoupling close-interaction instances.
The most successful model, BUCTD~\cite{BUCTD}, conditions top-down pose estimation with previously estimated keypoints from bottom-up methods.
It is a pose-refinement method with state-of-the-art results on OCHuman due to its strong ability to decouple closely interacting people.

\vspace{1em}
\noindent
\textbf{Human-Centered Foundational models.}
The latest directions in human body modeling are foundational models~\cite{Sapiens,HULK,UniHCP,DeepSortLab}.
They learn general features describing the human body that could be used for human-related tasks such as segmentation and pose estimation.
Most notably, Sapiens 2b~\cite{Sapiens} was trained on a staggering 2M images and, with 2B parameters, is almost four times bigger than ViTPose-h.
Even with this size, foundational models perform comparably or worse than much smaller specialized models. 

\newlength{\bmpcolw}
\setlength{\bmpcolw}{0.145\textwidth}

\NewDocumentCommand{\bmpimgnode}{m m m}{%
  \tikz[remember picture,baseline=(#1.base)]{
    \node[inner sep=0pt, outer sep=0pt] (#1) {\includegraphics[#2]{#3}};
  }%
}

\NewDocumentCommand{\bmpcolone}{m m m m m}{%
  \colorbox{#1}{%
    \begin{minipage}[t]{\bmpcolw}
      \centering
      \parbox[c][1.0em][c]{\linewidth}{\centering\small #2}\par\vspace{1pt}
      #3\par\vspace{2pt}
      #4\vspace{-2pt}
      {\footnotesize #5}
      \vspace{2pt}
    \end{minipage}%
  }%
}

\NewDocumentCommand{\bmpcoltwo}{m m m m m m m m}{%
  \colorbox{#1}{%
    \begin{minipage}[t]{2\bmpcolw}
      \centering
      \parbox[c][1.0em][c]{\linewidth}{\centering\small #2}\par\vspace{1pt}
      \begin{minipage}[t]{0.495\textwidth}
        \centering
        #3\par\vspace{2pt}
        #4
        {\footnotesize #5}
      \end{minipage}\hfill
      \begin{minipage}[t]{0.495\textwidth}
        \centering
        #6\par\vspace{2pt}
        #7
        {\footnotesize #8}
      \end{minipage}
    \end{minipage}%
  }%
}

\NewDocumentCommand{\bmparrow}{O{black} m m}{%
  \begin{tikzpicture}[remember picture,overlay]
    \draw[-{Latex[length=1.8mm,width=3.0mm]}, line width=3pt, draw=#1]
      ($(#2.east)!-0.80!(#2.north east)-(12pt,0)$)
      -- ($(#2.east)!-0.80!(#2.north east)+(-2pt,0)$);
  \end{tikzpicture}%
}

\NewDocumentCommand{\bmparrowLastLine}{O{black} m m}{%
  \begin{tikzpicture}[remember picture,overlay]
    \draw[-{Latex[length=1.8mm,width=3.0mm]}, line width=3pt, draw=#1]
      ($(#2.east)!-0.80!(#2.north east)-(7pt,0)$)
      -- ($(#2.east)!-0.80!(#2.north east)+(3pt,0)$);
  \end{tikzpicture}%
}
\NewDocumentCommand{\bmparrowFirstLine}{O{black} m m}{%
  \begin{tikzpicture}[remember picture,overlay]
    \draw[-{Latex[length=1.8mm,width=3.0mm]}, line width=3pt, draw=#1]
      ($(#2.west)!-0.80!(#2.north west)-(7pt,0)$)
      -- ($(#2.west)!-0.80!(#2.north west)+(3pt,0)$);
  \end{tikzpicture}%
}

\newlength{\bmptextcolw}
\newlength{\bmprowht}
\setlength{\bmptextcolw}{0.1\bmpcolw}
\setlength{\bmprowht}{1.0\bmpcolw}

\NewDocumentCommand{\bmptextcol}{m m}{%
  \begin{tabular}[c]{@{}c@{}}
    \rotatebox[origin=c]{90}{%
      \parbox[c][\bmptextcolw][c]{\bmprowht}{\centering #1}%
    }\\[2pt]
    \rotatebox[origin=c]{90}{%
      \parbox[c][\bmptextcolw][c]{\bmprowht}{\centering #2}%
    }%
  \end{tabular}%
}

\begin{figure*}[tb]
    \centering
    \raisebox{-15em}{\rotatebox{90}{\strut BMPv2 $2\times$ \hspace{3.2em} BMPv2 $1\times$}}%
    \hfill
    \bmpcolone{bmpColorMagenta!30}
        {Detector}
        {\bmpimgnode{det1}{width=0.99\textwidth}{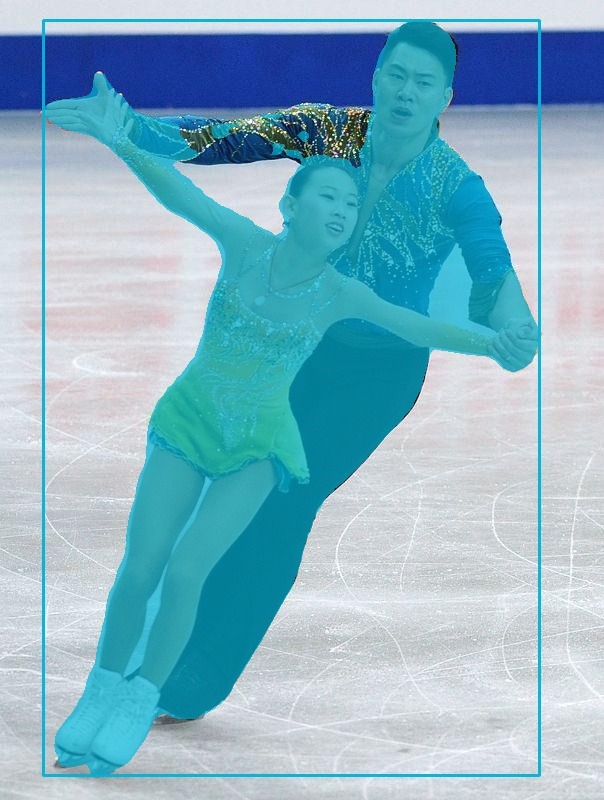}}
        {\bmpimgnode{det2}{width=0.99\textwidth}{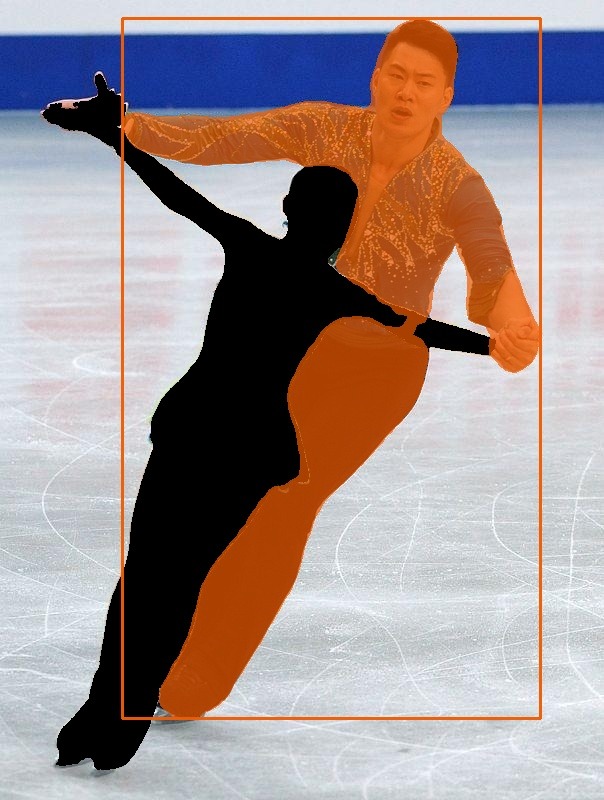}}
        {(A)}
\bmpcoltwo{bmpColorBlue!30}
        {PMPose}
        {\bmpimgnode{bck1}{width=\textwidth}{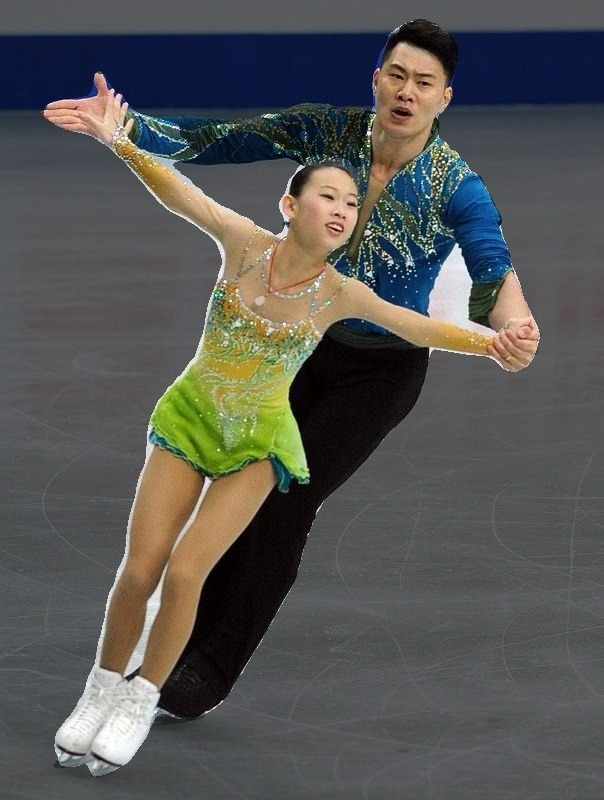}}
        {\bmpimgnode{bck2}{width=\textwidth}{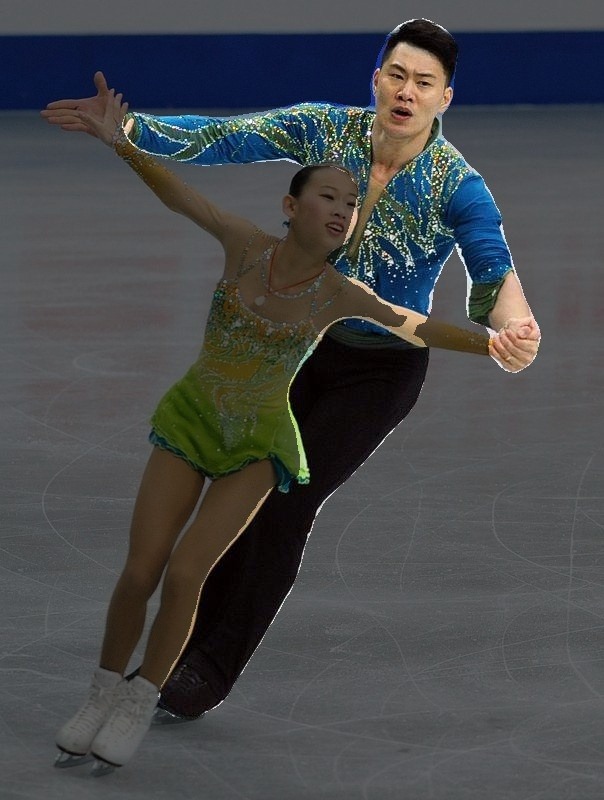}}
        {(B)}
        {\bmpimgnode{pose1}{width=\textwidth}{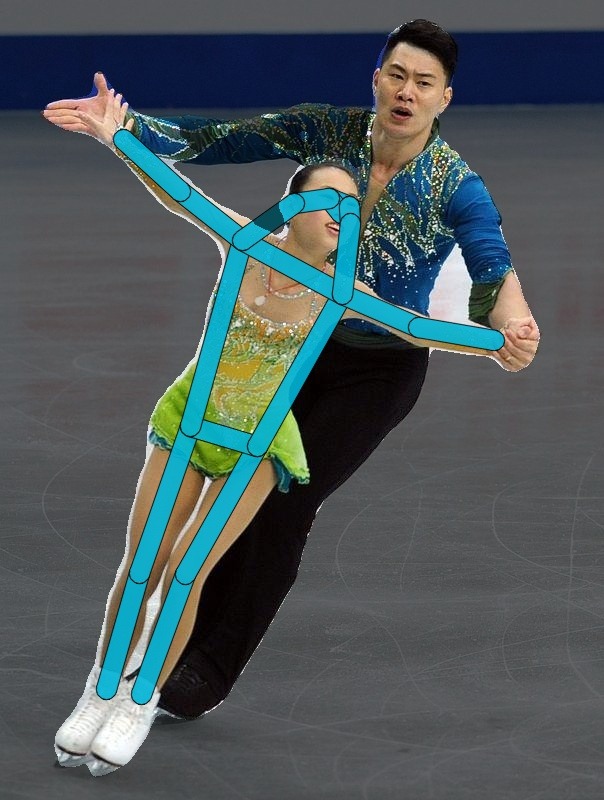}}
        {\bmpimgnode{pose2}{width=\textwidth}{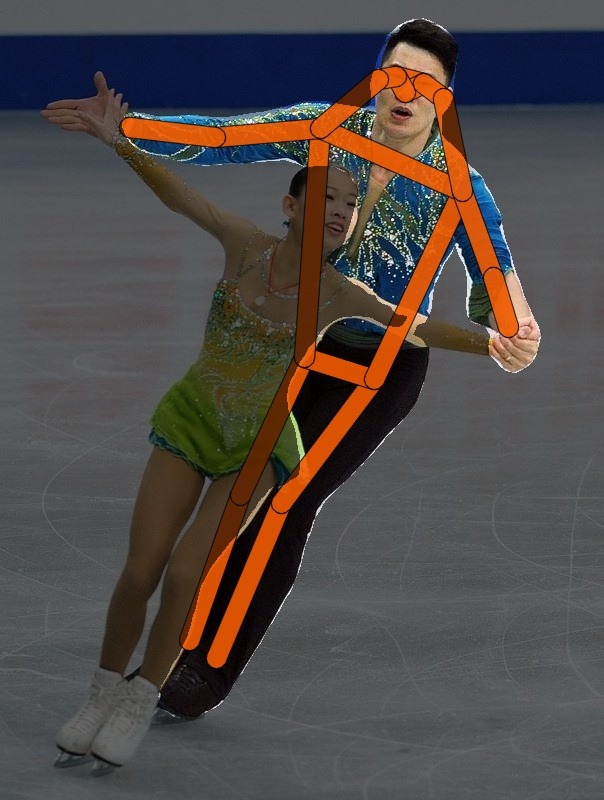}}
        {(C)}
\bmpcoltwo{bmpColorOrange!40}
        {SAM -- pose2seg}
        {\bmpimgnode{pts1}{width=\textwidth}{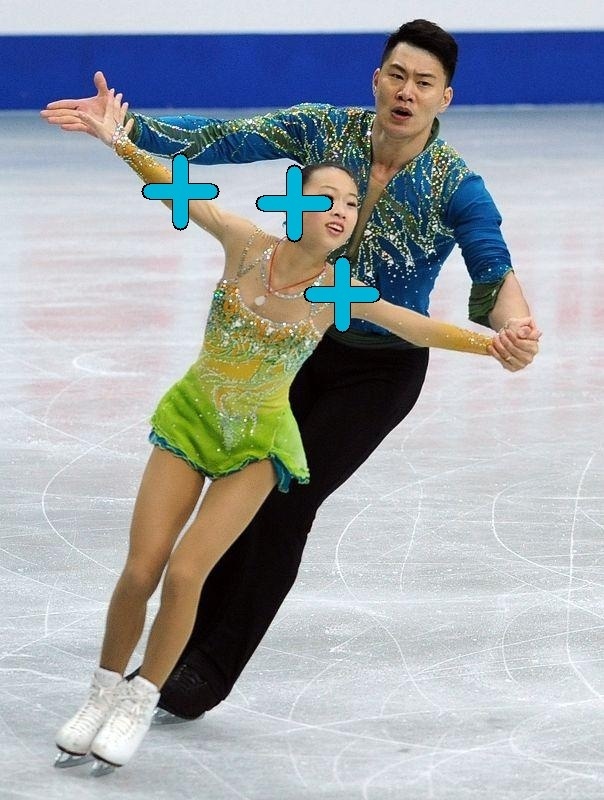}}
        {\bmpimgnode{pts2}{width=\textwidth}{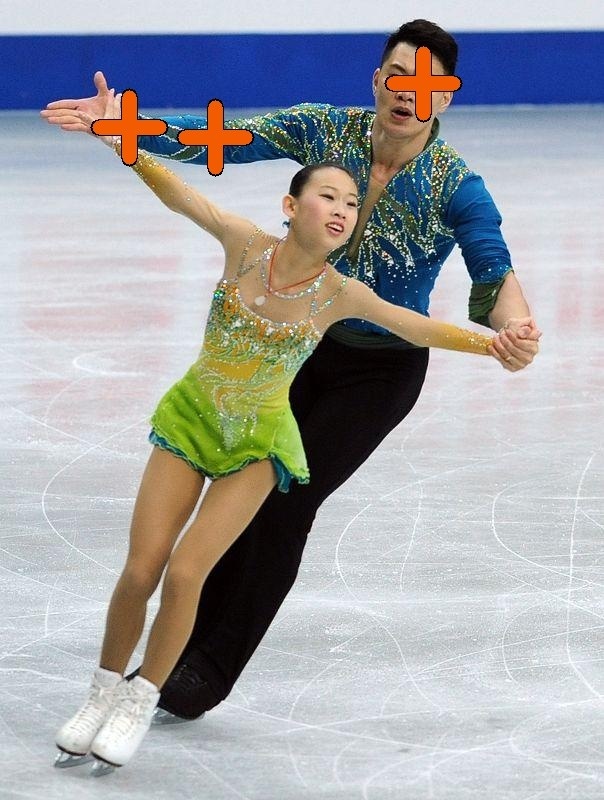}}
        {(D)}
        {\bmpimgnode{mask1}{width=\textwidth}{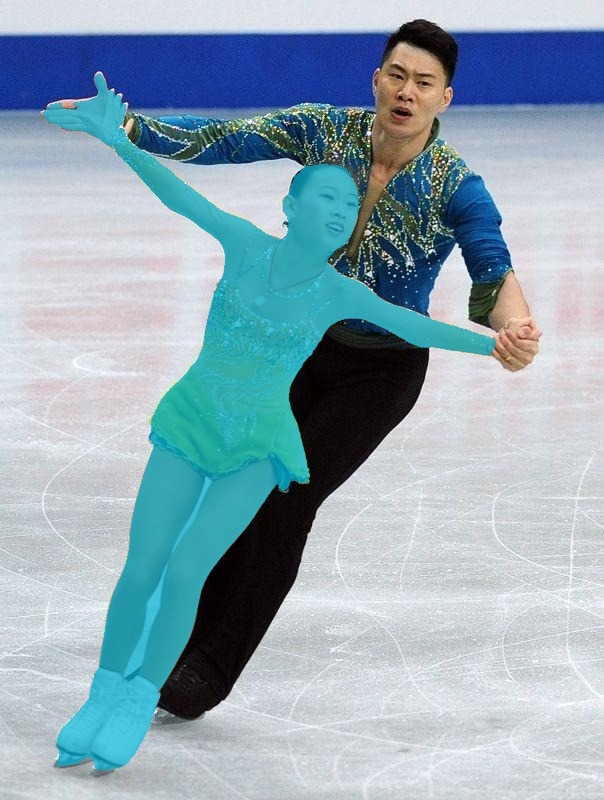}}
        {\bmpimgnode{mask2}{width=\textwidth}{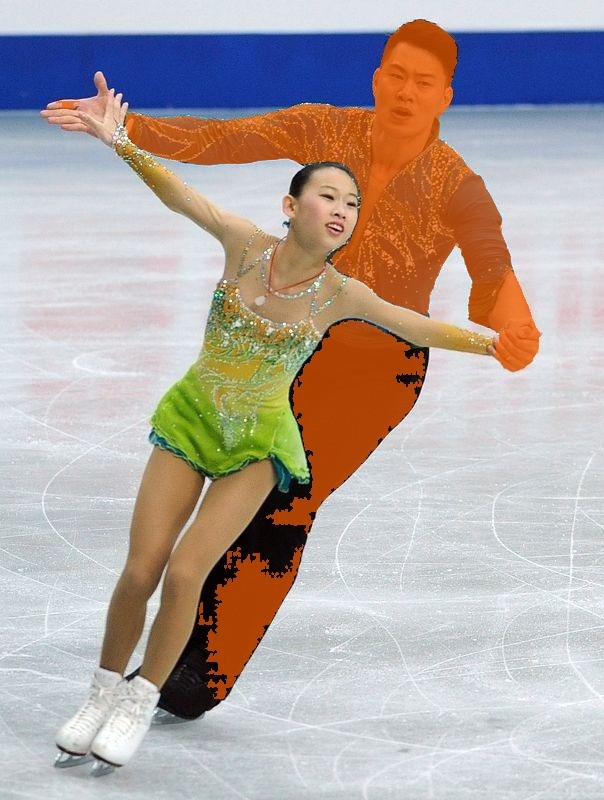}}
        {(E)}
\bmpcolone{white}
        {Black-out}
        {\bmpimgnode{black1}{width=0.99\textwidth}{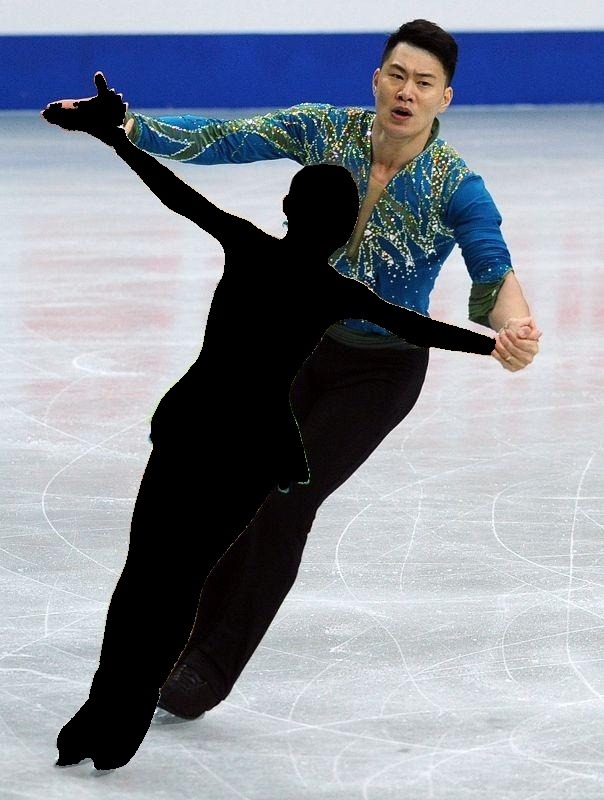}}
        {\bmpimgnode{black2}{width=0.99\textwidth}{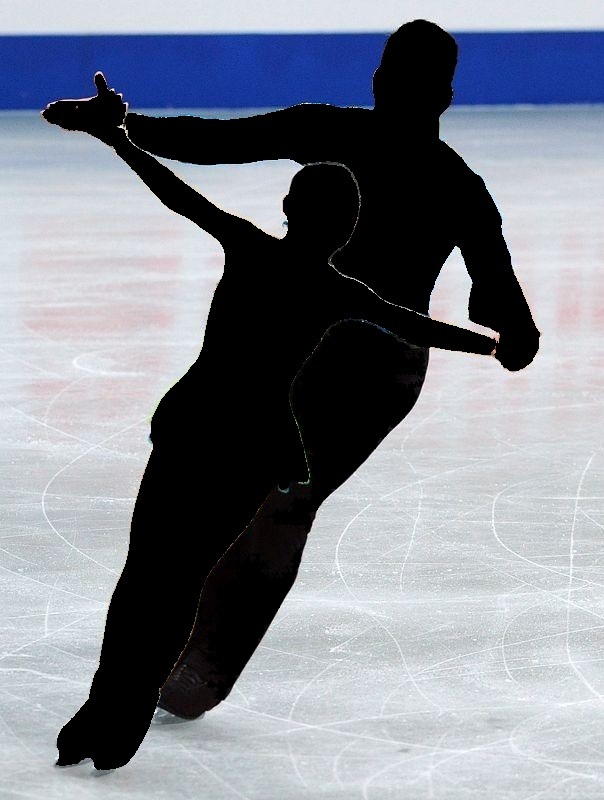}}
        {(F)}
    \bmparrow[bmpColorMagenta]{det1}{---}
    \bmparrow[bmpColorBlue]{bck1}{---}
    \bmparrow[bmpColorBlue]{pose1}{---}
    \bmparrow[bmpColorOrange]{pts1}{---}
    \bmparrow[bmpColorOrange]{mask1}{---}
    \bmparrowLastLine[black]{black1}{---}
    \bmparrowFirstLine[black]{det2}{---}
    \bmparrow[bmpColorMagenta]{det2}{--}
    \bmparrow[bmpColorBlue]{bck2}{---}
    \bmparrow[bmpColorBlue]{pose2}{---}
    \bmparrow[bmpColorOrange]{pts2}{---}
    \bmparrow[bmpColorOrange]{mask2}{---}
    \caption{
    \textbf{BBoxMaskPose v2 (BMPv2) loop visualization.}
    The method consists of three separate models --
    \colorbox{bmpColorMagenta!30}{detector}, \colorbox{bmpColorBlue!30}{PMPose} and \colorbox{bmpColorOrange!40}{SAM-pose2seg} --
    each conditioned by the others.
    A fine-tuned detector detects people in (A) while ignoring already blacked-out instances shown in (F).
    PMPose estimates pose (C) on person crops with dimmed background (B) to discriminate individuals in crowds.
    SAM-pose2seg refines masks (E) in full images conditioned by selected keypoints (D) to capture context.
    The loop ends after a fixed number of iterations or when no people are detected in (A).
    Image from OCHuman~\cite{pose2seg}.
    }
    \label{fig:BMP-visualization}
\end{figure*}

\vspace{1em}
\noindent
\textbf{Iterative Approaches:}
This work builds on the BBoxMaskPose loop~\cite{BMPv1}.
BMP combines three specialized models into a self-improving loop ideal for crowded and interaction-rich scenarios.
While BMP set a new SOTA on all three tasks (detection, segmentation, and 2D pose estimation), it still struggles in cases such as over-segmenting the human body or merging two individuals into the same pose.
Other iterative approaches are already mentioned BUCTD~\cite{BUCTD} and IterDet~\cite{IterDet}.
IterDet detects people in crowds iteratively, similarly to BMP.
This work improves BMPv1 with new pose estimation and pose-guided segmentation models and outperforms BUCTD and BMPv1 by a significant margin.

\vspace{1em}
\noindent
\textbf{3D Human Pose and Shape Estimation} has gained significant attention in recent years~\cite{HMRSurvey}.
Most methods rely on SMPL-based body models~\cite{SMPL, SMPL-X}, though several alternatives exist.
SMIL~\cite{SMIL} extends SMPL to infants.
ANNY~\cite{ANNY} models population-level variability across race, age, and body type.
SKEL~\cite{SKEL} augments SMPL with a realistic skeleton, while MHR~\cite{MHR} and ATLAS~\cite{ATLAS} model body shape directly around the physical skeleton.
While some models (e.g.~\cite{MultiHMR}) estimate 3D bodies directly from images, most 3D pose estimation methods require accurate 2D inputs.
Earlier approaches~\cite{NLF, METRO, GRAPHORMER} assume bounding-box–centered crops, whereas recent methods~\cite{PromptHMR, SAM-3D-Body} accept prompts such as text or segmentation masks.
BMPv2 provides precise and robust 2D predictions, making it well suited for 3D model prompting.

\vspace{1em}
\noindent
\textbf{Data:}
There are various datasets for 2D human pose estimation or instance segmentation.
The most notable are COCO~\cite{COCO}, MPII~\cite{MPII}, and AIC~\cite{AIC}.
Datasets such as OCHuman~\cite{pose2seg}, CrowdPose~\cite{CrowdPose}, and CIHP~\cite{CIHP} focus on multibody problems such as occlusion and self-occlusion.
OCHuman is too small for large-scale training and is traditionally used only for evaluation.
Furthermore, it contains annotation errors as shown in \cref{sec:data-OCH}.
CrowdPose is big enough for training but is unsuitable for evaluation in a multi-dataset setup as it mixes the train and test sets of COCO, MPII, and AIC.
In addition to the pose estimation, other datasets focus on human crowds, e.g. CrowdHuman~\cite{CrowdHuman} on detection or CIHP~\cite{CIHP} on human parsing.
3D human pose estimation datasets are mostly synthetic~\cite{BEDLAM, SURREAL, AGORA} or lab-captured~\cite{3DPW, h36m, 3DOH50k, CMUP}, so they do not evaluate performance in the wild.
Apart from COCO and OCHuman, there are not enough datasets to evaluate all three tasks (detection, segmentation and pose estimation) on the same dataset.
We attempt to improve the situation with OCHuman-Pose in \cref{sec:data-OCH}.
For COCO and related datasets, the evaluation metric is Object Keypoint Similarity (OKS), while Percentage of Correct Keypoints (PCKh) is used for MPII.

\section{Method}
\label{sec:method}

\begin{figure*}[tb]
    \centering

    \definecolor{method}{RGB}{196, 108, 44}    %
    \definecolor{famA}{RGB}{96, 150, 200}      %
    \definecolor{famB}{RGB}{120, 180, 130}     %
    \definecolor{famC}{RGB}{190, 170, 110}     %
    \definecolor{famF}{RGB}{145, 130, 185}     %
    \definecolor{famD}{RGB}{110, 175, 175}     %
    \definecolor{famE}{RGB}{205, 145, 165}     %
    \definecolor{otherA}{RGB}{165, 165, 165}   %
    \definecolor{otherB}{RGB}{115, 115, 115}   %

    \definecolor{vitcolor}{RGB}{230,126,34}      %
    \definecolor{pmcolor}{RGB}{40,100,210}       %
    \definecolor{hrformcolor}{RGB}{60,150,60}    %
    \definecolor{transcolor}{RGB}{200,170,0}     %
    \definecolor{cnncolor}{RGB}{150,80,200}      %
    \definecolor{mixcolor}{RGB}{20,150,160}      %
    \pgfplotsset{
        minimal/.style={
            axis line style={black!40},
            tick style={black!40},
            tick label style={font=\small},
            xlabel style={font=\small},
            ylabel style={font=\small},
            title style={font=\small},
            grid=both,
            grid style={black!5},
        }
    }   
    \begin{tikzpicture}
    \begin{axis}[
        minimal,
        width=\textwidth,
        height=9cm,
        axis y discontinuity=crunch, %
        axis x discontinuity=crunch, %
        xmin=34.2, xmax=50.5,
        ymin=64.9, ymax=83.6,
        xlabel={OCHuman test set AP (bboxes from RTMDet \cite{RTMDet})},
        ylabel={COCO val set AP (GT bboxes)},
        axis lines=left,
        grid=both,
        grid style=black!20,
        tick align=outside,
        enlargelimits=false,
        legend style={font=\small, at={(0.97,0.03)}, anchor=south east},
        legend cell align={left}
    ]
        \newcommand{\bubble}[5]{%
          \draw[fill=#4, draw=none, opacity=1.0]
            (axis cs:#1,#2) circle[radius=#3*4pt];
        }
        \newcommand{\MainBubble}[5]{%
          \draw[fill=#4, draw=black, opacity=1.0]
            (axis cs:#1,#2) circle[radius=#3*4pt];
        }
        \newcommand{\BUbubble}[5]{%
          \draw[dashed, draw=black!80, fill=#4, line width=0.5mm]
            (axis cs:#1,#2) circle[radius=#3*4pt];
        }

        \bubble{35.5}{75.9}{1.3}{famA}{S}
        \bubble{37.5}{77.9}{1.9}{famA}{B}
        \bubble{40.3}{80.7}{2.48}{famA}{L}
        \bubble{40.9}{81.4}{2.8}{famA}{H}
        \addplot[
            thick,
            famA,
            opacity=1.0,
        ] coordinates {
            (35.5, 75.9)
            (37.5, 77.9)
            (40.3, 80.7)
            (40.9, 81.4)
        };

        \bubble{44.0}{68.9}{1.48}{famB}{CID-w32}
        \bubble{45.0}{70.7}{1.8}{famB}{CID-w48}
        \addplot[
            thick,
            famB,
            opacity=1.0,
        ] coordinates {
            (44.0, 68.9)
            (45.0, 70.7)
        };

        \bubble{36.5}{71.0}{1.8}{famF}{DEKR}
        
        \bubble{41.3}{66.1}{2.47}{famF}{Sapiens 0.3b}
        \bubble{40.0}{69.5}{1.7}{famF}{HQNetR-50}
        \bubble{42.5}{76.3}{1.7}{famF}{MIPNet}
        \bubble{47.4}{74.8}{2.0}{famF}{BUCTD}
        
        \MainBubble{46.78}{77.85}{1.36}{vitcolor}{S} %
        \MainBubble{48.2}{79.4}{1.95}{vitcolor}{B}
        \MainBubble{49.3}{80.31}{2.49}{vitcolor}{L}
        \MainBubble{49.9}{80.7}{2.8}{vitcolor}{H}  %
        \addplot[
            thick,
            vitcolor,
            opacity=1.0,
        ] coordinates {
            (46.78, 77.85)
            (48.2, 79.4)
            (49.3, 80.31)
            (49.9, 80.7)  %
        };

        \MainBubble{55.3}{79.4}{3.0}{red}{B}

        \bubble{36.3}{70.1}{0.68}{famD}{T}
        \bubble{39.0}{74.1}{0.95}{famD}{S}
        \bubble{40.2}{78.0}{1.39}{famD}{M}
        \bubble{40.8}{78.6}{1.72}{famD}{L}
        \addplot[
            thick,
            famD,
            opacity=1.0,
        ] coordinates {
            (36.3, 70.1)
            (39.0, 74.1)
            (40.2, 78.0)
            (40.8, 78.6)
        };

        \bubble{38.5}{76.7}{1.46}{famE}{w32}
        \bubble{39.3}{77.3}{1.8}{famE}{w48}
        \addplot[
            thick,
            famE,
            opacity=1.0,
        ] coordinates {
            (38.5, 76.7)
            (39.3, 77.3)
        };

    \node[font=\normalsize, anchor=center, rotate=20]
            at (axis cs:38.2, 75.3) {\textbf{\textcolor{famE}{{\hypersetup{citecolor=famE}HRNet \cite{HRNet}}}}};
    \node[font=\scriptsize, anchor=center, rotate=0]
            at (axis cs:38.5, 77.6) {\textbf{\textcolor{famE}{w32}}};
    \node[font=\scriptsize, anchor=center, rotate=0]
            at (axis cs:39.3, 78.5) {\textbf{\textcolor{famE}{w48}}};
            
    \node[font=\normalsize, anchor=center, rotate=20]
            at (axis cs:38.0, 80.8) {\textbf{\textcolor{famA}{ViTPose {\hypersetup{citecolor=famA}\cite{ViTPose}}}}};
    \node[font=\scriptsize, anchor=center, rotate=0]
            at (axis cs:35.5, 76.8) {\textbf{\textcolor{famA}{S}}};
    \node[font=\scriptsize, anchor=center, rotate=0]
            at (axis cs:37.5, 79.0) {\textbf{\textcolor{famA}{B}}};
    \node[font=\scriptsize, anchor=center, rotate=0]
            at (axis cs:40.3, 82.0) {\textbf{\textcolor{famA}{L}}};
    \node[font=\scriptsize, anchor=center, rotate=0]
            at (axis cs:40.9, 82.9) {\textbf{\textcolor{famA}{H}}};
            
    \node[font=\normalsize, anchor=center, rotate=30]
            at (axis cs:41.2, 76.8) {\textbf{\textcolor{famD}{RTMPose {\hypersetup{citecolor=famD}\cite{RTMPose}}}}};
    \node[font=\scriptsize, anchor=center, rotate=0]
            at (axis cs:36.3, 69.5) {\textbf{\textcolor{famD}{T}}};
    \node[font=\scriptsize, anchor=center, rotate=0]
            at (axis cs:39.0, 73.2) {\textbf{\textcolor{famD}{S}}};
    \node[font=\scriptsize, anchor=center, rotate=0]
            at (axis cs:40.2, 77.0) {\textbf{\textcolor{famD}{M}}};
    \node[font=\scriptsize, anchor=center, rotate=0]
            at (axis cs:40.8, 77.5) {\textbf{\textcolor{famD}{L}}};

    \node[font=\normalsize, anchor=center, rotate=38]
            at (axis cs:44.99, 69.0) {\textbf{\textcolor{famB}{{\hypersetup{citecolor=famB}CID \cite{CID}}}}};
    \node[font=\scriptsize, anchor=center, rotate=0]
            at (axis cs:44.0, 70.0) {\textbf{\textcolor{famB}{w32}}};
    \node[font=\scriptsize, anchor=center, rotate=0]
            at (axis cs:45.0, 72.0) {\textbf{\textcolor{famB}{w48}}};

    \node[font=\normalsize, anchor=center, rotate=0]
            at (axis cs:36.5, 72.2) {\textbf{\textcolor{famF}{{\hypersetup{citecolor=famF}DEKR \cite{DEKR}}}}};
    \node[font=\normalsize, anchor=center, rotate=0]
            at (axis cs:40.0, 70.8) {\textbf{\textcolor{famF}{{\hypersetup{citecolor=famF}HQNetR-50 \cite{HQNet}}}}};
    \node[font=\normalsize, anchor=center, rotate=0]
            at (axis cs:42.5, 75.2) {\textbf{\textcolor{famF}{MIPNet {\hypersetup{citecolor=famF}\cite{MIPNet}}}}};
    \node[font=\normalsize, anchor=center, rotate=0]
            at (axis cs:47.4, 73.4) {\textbf{\textcolor{famF}{{\hypersetup{citecolor=famF}BUCTD \cite{BUCTD}}}}};
    \node[font=\normalsize, anchor=center, rotate=0]
            at (axis cs:41.3, 67.5) {\textbf{\textcolor{famF}{Sapiens 0.3b {\hypersetup{citecolor=famF}\cite{Sapiens}}}}};
    
    \node[font=\large, anchor=center, rotate=20]
            at (axis cs:47.5, 81.8) {\textbf{\textcolor{vitcolor}{PMPose}}};
    \node[font=\normalsize, anchor=center, rotate=0]
            at (axis cs:46.78, 78.85) {\textbf{\textcolor{vitcolor}{S}}};
    \node[font=\normalsize, anchor=center, rotate=0]
            at (axis cs:48.2, 80.7) {\textbf{\textcolor{vitcolor}{B}}};
    \node[font=\normalsize, anchor=center, rotate=0]
            at (axis cs:49.3, 81.8) {\textbf{\textcolor{vitcolor}{L}}};
    \node[font=\normalsize, anchor=center, rotate=0]
            at (axis cs:49.9, 82.3) {\textbf{\textcolor{vitcolor}{H}}};

    \end{axis}
    \end{tikzpicture}

    \caption{
    \textbf{%
    Comparison of
    \textcolor{vitcolor}{PMPose}
    with state-of-the-art models} on COCO and OCHuman.
    The bubble diameter is proportional to the logarithm of the number of model parameters.
    PMPose generalizes best to the unseen OCHuman dataset while keeping a SOTA-level performance on COCO.
    }
    \label{fig:PMPose-vs-ViTPose}
\end{figure*}

BMPv1~\cite{BMPv1} connects a detector, pose estimator, and segmenter in a self-improving loop as shown in \cref{fig:BMP-visualization}.
The core idea is mutual consistency between different human-body representations.
Each model is conditioned on the others and iteratively refines its predictions.
The loop ends when no people are detected.

The detector is conditioned on previously processed instances.
After an instance is processed, its mask is blacked out, forcing the detector to ignore it in subsequent iterations.
This enables discovery of missed instances, typically in the background, and recovers cases where multiple people collapse into a single detection.

The pose estimator is conditioned on a segmentation mask from the detector or the mask refinement stage.
Masks are more precise than bounding boxes and help disentangle individuals under heavy overlap.
The pose estimator in BMPv1, MaskPose, is a standard heatmap-based top-down model with mask conditioning.

The final component of BMPv1 is SAM, which refines segmentation masks using previous estimates and predicted poses.
BMPv1 proposes a complex strategy for selecting pose-based point prompts.
Despite this, mask refinement remains the dominant failure mode.

In this work, we revisit BMP and introduce two improvements.
We replace MaskPose with PMPose and SAM2 with SAM-pose2seg.
Both are introduced in the following sections.
We further introduce a more expensive but more accurate variant, BMP+, and show how BMP integrates with 3D pose estimation.

\subsection{PMPose}
\label{sec:method-PMPose}

Human pose estimation is a central element of the BMP loop.
It resolves detection ambiguity in crowds by enforcing anatomical constraints (e.g., each person has two arms) and validates detections before mask refinement.
Accurate poses are critical for the final refinement stage; incorrect joints prevent the separation of overlapping individuals.

The dominant failure modes are (1) merging people when predicted joints belong to multiple individuals and (2) missing disconnected parts, such as the wrists or ankles that are visually detached due to occlusion.
BMPv1 uses MaskPose~\cite{BMPv1}, which benefits from mask conditioning and reduces merges between individuals.
However, MaskPose inherits limitations common in top-down pose estimators.
First, because it is trained only on in-image keypoints, it tends to place joints inside the image even when they lie outside, causing instabilities along the bounding box border.
Second, its confidence estimates are poorly calibrated.
Confidence is crucial for selecting prompts in the mask refinement stage, and overconfident errors lead to incorrect prompt selection that propagates through the pipeline.
We address these issues with PMPose (ProbMaskPose), a unified model for pose estimation in BMPv2.

PMPose combines the strengths of MaskPose and ProbPose~\cite{ProbPose}.
It keeps mask conditioning from MaskPose and adopts the presence probability, visibility estimation, expected OKS prediction, and the resulting border stability from ProbPose.
The ViT backbone remains unchanged; only the prediction head and the loss differ.

PMPose follows the same COCO+AIC+MPII training setup as MaskPose and is further fine-tuned with CropAugmentation on the same datasets.
CropAugmentation enables the localization of keypoints outside the visible region and improves the presence and visibility predictions.
For the semi-transparent visibility masking introduced by MaskPose, we use $\alpha = 0.25$ (compared to $0.2$ in MaskPose), which is bit-shift friendly and speeds up augmentation.

PMPose has the same outputs as ProbPose~\cite{ProbPose}: keypoints with calibrated probability maps ($\mathcal{K}$), presence probability ($p_p$) indicating whether each joint is present, visibility estimates ($v$), and expected OKS ($\mathbb{E}_{\mathit{OKS}}$) as a self-quality measure.
All of them are conditioned on an instance segmentation mask and bounding box as in MaskPose~\cite{BMPv1}.
In the BMP loop, visibility is used to select prompts in the pose refinement stage (see \cref{tab:sam-pose2seg-prompt}) and $\mathbb{E}_{\mathit{OKS}}$ is used as a confidence measure for COCO-style AP evaluation.

Probability maps are calibrated with the same process (temperature scaling) as in ProbPose.
We quantize the probabilities in 5\% increments and ensure each range contains approximately 5\% ground truth keypoints of the CropCOCO validation set.

We introduce a whole family of models: PMPose-S/B/L/H, paired with ViT-S/B/L/H backbones~\cite{ViT}.
The design is consistent across scales; capacity is the only difference.
PMPose outperforms ViTPose~\cite{ViTPose} and improves over MaskPose, especially in crowded scenes (\cref{fig:PMPose-vs-ViTPose}).
On COCO, performance matches state-of-the-art levels.
PMPose-S provides practical speed for real-time or near-real-time use, while PMPose-H targets the highest accuracy.

\subsection{SAM-pose2seg}
\label{sec:method-SAM}

SAM-pose2seg builds on SAM 2.1~\cite{SAM2} and we first review the baseline to clarify our modifications.

SAM is trained on a large and diverse dataset, which gives it strong generalization to unseen data.
However, SAM lacks explicit semantic understanding.
Objects are loosely defined and may correspond to almost arbitrary regions.

Our goal is to preserve SAM’s generalization while specializing it for pose-guided human instance segmentation.
We therefore introduce minimal changes and keep most of the original architecture unchanged.

During training, SAM uses point prompts dynamically sampled from the ground-truth mask.
For each mask, a single point is uniformly sampled and used to predict a mask.
To simulate refinement, additional points are sampled from the error region, defined as the area where the prediction disagrees with the ground truth.
This process is repeated for eight iterations, which results in the behavior observed in BMPv1~\cite{BMPv1}: increasing the number of prompt points can degrade performance rather than improve it.

During both training and inference, SAM predicts three candidate masks per prompt.
A mask is selected based on the predicted confidence score.
This design partially addresses object ambiguity that is not explicitly resolved during the original training procedure.

\subsubsection{Adapting SAM for the pose2seg task}

We fine-tune from the original SAM 2.1.
The backbone is frozen to preserve generalization.
The architecture remains unchanged; only the training procedure is modified.

We adapt the model in two ways.

\noindent
\textbf{Human-focused decoder fine-tuning}.
We fine-tune the decoder to predict only human instances, including clothing as defined by the training annotations.
This removes task ambiguity and directly addresses the main failure mode of BMPv1: over-segmentation of skin regions.

\noindent
\textbf{Pose-guided prompting during training.}
We replace random point prompting with pose-aware prompting to inject human keypoint information into the model.
The first prompt is the most visible keypoint.
In subsequent iterations, we select a keypoint located in the current error region.
If no keypoint lies in the error region, we fall back to sampling a random point from that region.
As a result, the model learns to operate on pose-driven prompts, which have a different spatial distribution than the random points used in the original SAM training.

This sampling strategy incorporates pose cues directly into training and better aligns the model with pose-guided human segmentation.
Training uses human keypoints predicted with ProbPose~\cite{ProbPose}, allowing the use of datasets without keypoint ground truth (CIHP~\cite{CIHP}).

Fine-tuning also enables rapid specialization to the human class.
Consequently, the complex prompting strategy identified as optimal in BMPv1 is no longer necessary.
Selecting a small number of high-visibility keypoints yields comparable or better performance while being significantly simpler.
Experiments show that two keypoints are sufficient to characterize an instance; we use three keypoints in the final model for slightly improved performance.

\cref{sec:exp} analyzes how SAM-pose2seg improves the BMP loop and its connection to PMPose through visibility and presence probability prediction.
Additional ablations are provided in the technical report~\cite{SAM-pose2seg-report}.

\subsection{BMP+}
\label{sec:bmp-plus}

BMPv1~\cite{BMPv1} ablations showed that re-running the pose estimator after mask refinement brought only marginal gains, not justifying the extra computation.
This no longer holds for BMPv2.
With stronger components—SAM-pose2seg for masks and PMPose for pose—looping them yields significant gains.
We therefore propose BMPv2+, a higher-performance, higher-cost variant of BMPv2.

Like BMPv2, BMPv2+ starts with instance detection, followed by pose estimation and mask refinement.
The key difference is that pose estimation and mask refinement are looped until convergence.
In practice, one additional PMPose pass is sufficient, as the second SAM-pose2seg step brings no further gains.

This loop increases computational cost but improves accuracy, particularly for human pose estimation. BMPv2+ is suited for use cases where performance outweighs runtime and resource constraints.

\subsection{From 2D to 3D}
\label{sec:method-3D}

BMPv1~\cite{BMPv1} demonstrated strong robustness in crowded scenes.
In BMPv2, we extend this capability to 3D by leveraging reliable 2D masks, boxes, and poses to reason about human bodies under dense interactions.
Our goal is to recover accurate 3D poses and shapes despite extreme overlap and occlusion.

BMPv2 uses masks, bboxes, and 2D poses as prompts for a 3D human pose estimator.
The 3D step is executed only after the BMP loop converges, as 3D models are computationally heavy.
Although 3D estimation could be integrated into the loop similarly to the 2D pose, its cost would make iterative refinement impractical.

The 3D pose predictor encodes the full image once to obtain the image features.
For each detected individual, it predicts 3D pose and shape by prompting with the estimated instance’s bbox and mask.
This avoids repeated feature extraction and keeps the per-person overhead manageable.

Predicted 3D bodies are filtered using 3D non-maximum suppression.
Instance confidence for NMS is inherited from the 2D predictions produced by the BMP loop.

Quantitative results are in \cref{sec:exp-3D}, with qualitative examples in \cref{fig:teaser}.

\section{Data}
\label{sec:data}

BMP targets scenes with very close human interactions where bounding boxes alone cannot separate individuals.
Evaluating such models is difficult due to the lack of suitable data.
To our knowledge, only CrowdPose~\cite{CrowdPose} and OCHuman~\cite{pose2seg} focus on human pose estimation in such scenes.
CrowdPose, however, is problematic in multi-dataset training due to data leakage, while \cref{fig:PMPose-vs-ViTPose} shows that multi-dataset training is essential for strong generalization.
This leaves OCHuman as the only viable option, yet we show below that it is insufficient for in-the-wild detection and pose estimation.

The following two sections analyze the OCHuman and CIHP datasets, their statistics, how they differ, and what their weaknesses are.
We show that annotations in OCHuman are incomplete and, therefore, that detection results on the dataset are misleading.
To address this gap, we introduce an improved variant of OCHuman, OCHuman-Pose.
The dataset is available on the \ProjectWebsite.

\subsection{OCHuman-Pose}
\label{sec:data-OCH}

\begin{table}[b]
\newcommand{\overbar}[1]{\mkern 1.5mu\overline{\mkern-1.5mu#1\mkern-1.5mu}\mkern 1.5mu}
\newcolumntype{C}{>{\centering\arraybackslash}X}
\caption{
\textbf{The ``crowded'' datasets:} OCHuman, CIHP and the new OCHuman-Pose.
$\overbar{\IoUMax}$ is the mean of $\IoUMax$ over all instances.
In OCHuman-Pose, more than 50\% of previously un-annotated instances are added to OCHuman images.
CIHP is less crowded (see \cref{fig:OCH-histograms}) but larger and its masks are more precise. 
}
\label{tab:ochuman-original-stats}%
\begin{tabularx}{\linewidth}{@{}l|CC|CC|C@{}}
\toprule
            & \multicolumn{2}{c}{OCH} & \multicolumn{2}{c}{OCH-Pose} & CIHP \\
             & val   & test & val & test & val \\
\midrule
\footnotesize\#\scriptsize\hspace{0.3em}imgs      & 2 500  & 2 231   & 2 500  & 2 231 & 5 000 \\
\footnotesize\#\scriptsize\hspace{0.3em}kpts & 4 291  & 3 819   & 6 546  & 5 863  & 0  \\
\footnotesize\#\scriptsize\hspace{0.3em}masks & 4 291  & 3 819   & 4 291  & 3 819  & 17 520 \\
$\overbar{\IoUMax}$     & 0.545   & 0.546    & 0.527   & 0.527 & 0.209  \\
\botrule
\end{tabularx}
\vspace{-1.5em}
\end{table}

\begin{figure}[tb]
\centering

\pgfmathsetmacro{\meanA}{0.545}   %
\pgfmathsetmacro{\meanB}{0.527}

\pgfplotsset{
    minimal/.style={
        ybar,
        ymin=0,
        bar width=3.5pt,              %
        axis line style={black!40},
        tick style={black!40},
        tick label style={font=\small},
        xlabel style={font=\small},
        ylabel style={font=\small},
        title style={font=\small},
        grid=both,
        grid style={black!20},
    }
}

\begin{tikzpicture}

\begin{axis}[
    minimal,
    width=\linewidth,
    height=0.8\linewidth,
    xlabel={$\IoUMax$},
    ylabel={\% of instances},
    xmin=-0.01, xmax=1.01,
    ymin=0, ymax=0.23,
    xtick={0,0.2,...,1.00},
    ytick={0,0.1,...,0.2},
    legend style={
        font=\small,
        at={(0.98,0.98)},
        anchor=north east,
        nodes={anchor=west},
    },
    legend cell align={left},
    legend image code/.code={
        \draw[draw=none] (0,0) rectangle (0,0);
    },
]

\pgfmathsetmacro{\barW}{2.5}

\addplot+[
    draw=none,
    fill=bmpColorOrange,
    fill opacity=1.0,
    bar shift=-\barW pt,
    bar width=\barW pt,
] coordinates {
    (0.025,0.2198) (0.075,0.1229) (0.125,0.1153) (0.175,0.1068) (0.225,0.0879)
    (0.275,0.0746) (0.325,0.0580) (0.375,0.0535) (0.425,0.0465) (0.475,0.0318)
    (0.525,0.0243) (0.575,0.0179) (0.625,0.0136) (0.675,0.0089) (0.725,0.00736)
    (0.775,0.00417) (0.825,0.00263) (0.875,0.00274) (0.925,0.00080) (0.975,0.00057)
};

\addplot+[
    draw=none,
    fill=bmpColorMagenta!70,
    fill opacity=1.0,
    bar shift=0pt,
    bar width=\barW pt,
] coordinates {
    (0.025,0.1849) (0.075,0.00052) (0.125,0.00052) (0.175,0.0) (0.225,0.00079)
    (0.275,0.00157) (0.325,0.0) (0.375,0.00026) (0.425,0.0) (0.475,0.0)
    (0.525,0.1709) (0.575,0.1333) (0.625,0.0990) (0.675,0.0757) (0.725,0.0550)
    (0.775,0.1241) (0.825,0.0807) (0.875,0.0429) (0.925,0.0204) (0.975,0.0094)
};

\addplot+[
    draw=none,
    fill=bmpColorBlue!80,
    fill opacity=1.0,
    bar shift=+\barW pt,
    bar width=\barW pt,
] coordinates {
    (0.025,0.0394) (0.075,0.0244) (0.125,0.0203) (0.175,0.0258) (0.225,0.0283)
    (0.275,0.0403) (0.325,0.0449) (0.375,0.0403) (0.425,0.0456) (0.475,0.0384)
    (0.525,0.1424) (0.575,0.1090) (0.625,0.0878) (0.675,0.0706) (0.725,0.0476)
    (0.775,0.0884) (0.825,0.0553) (0.875,0.0297) (0.925,0.0154) (0.975,0.00648)
};

\legend{
    {\color{bmpColorOrange}\rule{5pt}{5pt}}~CIHP,
    {\color{bmpColorMagenta!70}\rule{5pt}{5pt}}~OCH 1.0,
    {\color{bmpColorBlue!80}\rule{5pt}{5pt}}~OCH-Pose
}

\end{axis}
\end{tikzpicture}

\caption{
\textbf{Instance density} measured by $\IoUMax$ in OCHuman, OCHuman-Pose and CIHP datasets.
Despite having almost the same mean of $\IoUMax$ as OCHuman (\cref{tab:ochuman-original-stats}), OCHuman-Pose has a more realistic distribution of in-the-wild images, free from artificial instance selections.
CIHP has a different distribution because it focuses on crowds instead of close human interactions.
}
\label{fig:OCH-histograms}
\end{figure}

\begin{figure}[tb]
    \centering
    \begin{subfigure}{0.305\linewidth}
        \centering
        \includegraphics[width=\textwidth]{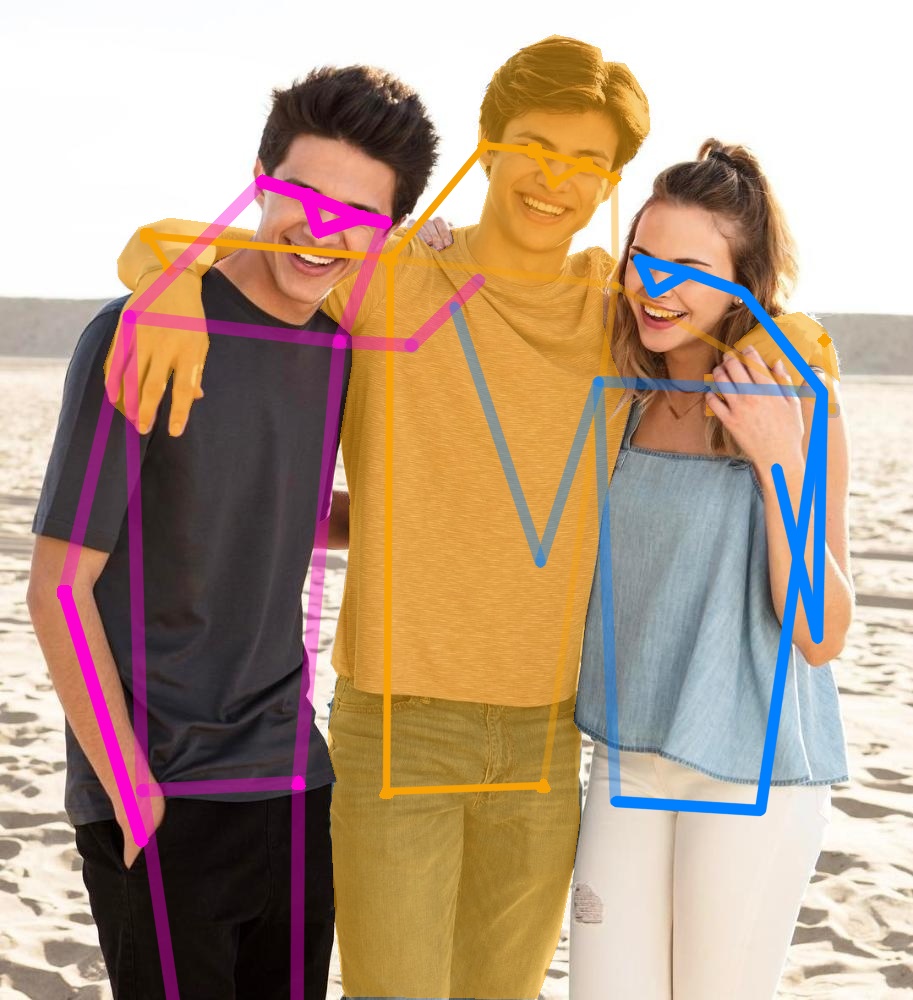}
    \end{subfigure}
    \hfill
    \begin{subfigure}{0.257\linewidth}
        \centering
        \includegraphics[width=\textwidth]{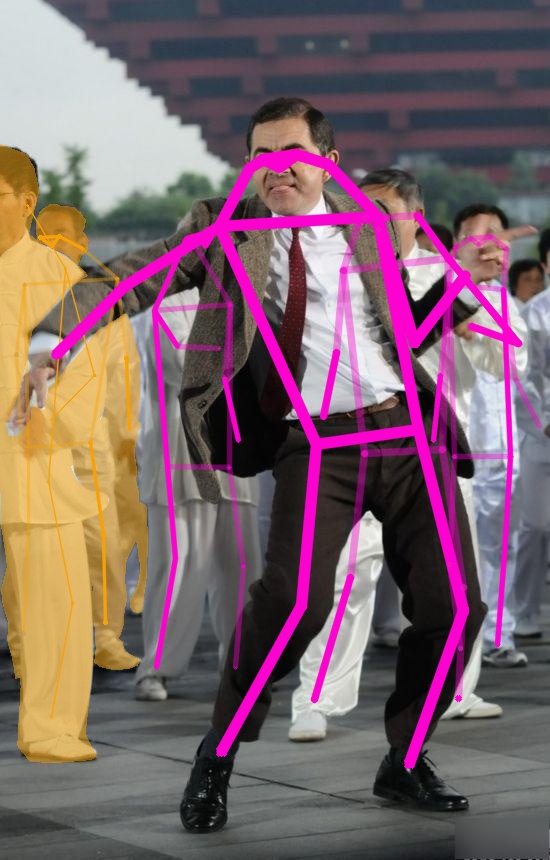}
    \end{subfigure}
    \hfill
    \begin{subfigure}{0.40\linewidth}
        \centering
        \includegraphics[width=\textwidth]{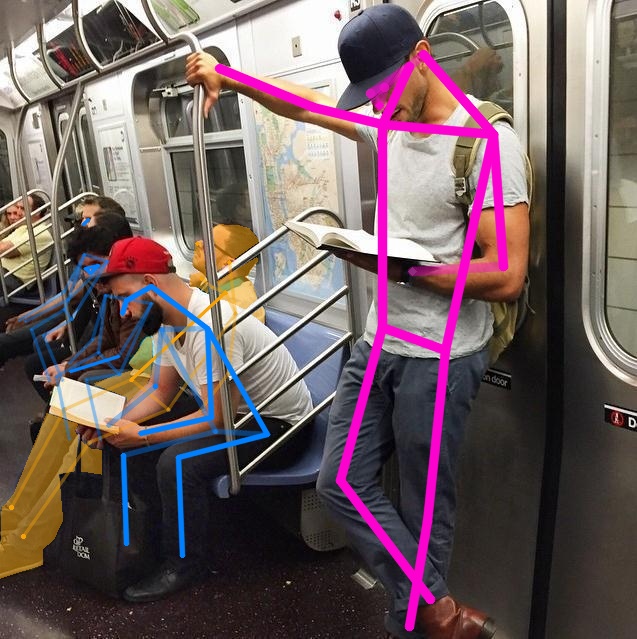}
    \end{subfigure}
    \caption{
    \textbf{Missing annotations} in OCHuman.
    \textcolor{bmpColorOrange}{The~originally annotated instances} are often in the background, which causes problems in COCO-style AP evaluation.
    OCHuman-Pose expands the original dataset with \textcolor{bmpColorMagenta}{new instances} and \textcolor{bmpColorBlue}{previously ignored annotations}.    
    \vspace{-1.5em}
    }
    \label{fig:OCH-missing-annots}
\end{figure}

\begin{table}[b]
    \centering
    \newcolumntype{C}{>{\centering\arraybackslash}X}
    \caption{
    \textbf{ViTPose*-b}~\cite{ViTPose} \textbf{performance}
    on OCHuman and new OCHuman-Pose using 
    GT bounding boxes (top) and RTMDet~\cite{RTMDet} (bottom) bounding boxes as input.
    {Missing annotations in OCHuman are evaluated as detection errors (false positives), masking pose estimation errors.}
    On~OCHuman-Pose, performance with GT and detected bboxes is much closer, which aligns with the qualitative analysis.
    }
    \label{tab:och-pose-gt-vs-det}
    \begin{tabularx}{\linewidth}{l| CC | Cc}
    \toprule
    GT      & \multicolumn{2}{c|}{OCHuman} & \multicolumn{2}{c}{OCHuman-Pose} \\ 
    bboxes  &  val AP & test AP           & val AP & test AP \\
    \midrule
    \cmark    &  90.9   & 91.0              & 86.4   & 86.2    \\
    \xmark    &  44.5   & 44.1              & 75.3   & 76.1    \\
    \bottomrule
    \end{tabularx}
    \vspace{-1.5em}
\end{table}

OCHuman was originally created for the pose-to-segmentation task, where the goal is to estimate an instance mask given a human pose.
Although not designed for general-purpose pose estimation, it became the de-facto benchmark for in-the-wild crowded human pose estimation due to the absence of better alternatives.

The main issue with OCHuman is that many individuals are not annotated.
For each person, the dataset defines $\IoUMax$ as the maximum IoU with any other instance in the image.
The authors kept only cases with $\IoUMax=0$ or $\IoUMax>0.5$, excluding all instances with intermediate overlap ($0 < \IoUMax \le 0.5$), as shown in \cref{fig:OCH-histograms}.
This filtering leaves numerous unannotated people, often placing the annotated subjects in the background, as shown in \cref{fig:OCH-missing-annots}.

The COCO-style mAP evaluation~\cite{COCO} sorts the predictions by confidence and assigns the ground truth greedily.
Any prediction that ranks first without a matching annotation becomes a false positive.
On OCHuman, this forces models to assign the highest confidence to the annotated instance, even when it is not the most prominent person in the image.
As illustrated in \cref{fig:OCH-missing-annots}, there is no reason for these annotated individuals -- often in the background -- to receive the highest confidence, yet the evaluation protocol penalizes any alternative.

We address this issue by annotating all missing instances with keypoints.
As shown in \cref{fig:OCH-missing-annots}, some individuals had keypoints in the original OCHuman annotations but were excluded due to the dataset’s filtering scheme.
We reinstate these annotations and add manual annotations for all instances that lacked them.

\cref{tab:ochuman-original-stats} summarizes the statistics of the revised dataset.
OCHuman-Pose adds over 50\% new instances to the original images.
Although the average $\IoUMax$ remains similar, the distribution now better reflects in-the-wild conditions (\cref{fig:OCH-histograms}): the number of isolated cases with $\IoUMax=0$ is reduced and the previously missing range $0 < \IoUMax \le 0.5$ is now represented.
Measuring pose estimation performance on OCHuman-Pose reflects reality more accurately than the original OCHuman as shown in \cref{tab:och-pose-gt-vs-det}.
Annotation details are described at Suppl. on the \ProjectWebsite.

Note that OCHuman-Pose does not include masks for the newly added persons, and thus cannot be used to evaluate segmentation mAP.

\subsection{CIHP}
\label{sec:data-CIHP}

Because OCHuman-Pose lacks masks and the original OCHuman annotates only a subset of instances, we require an additional dataset for evaluating instance segmentation in crowded scenes.
CIHP contains dense crowds rather than close-contact interactions, leading to different bounding-box overlap patterns (see $\IoUMax$ distribution in \cref{fig:OCH-histograms}).
It is substantially larger than OCHuman and provides more precise masks, but it has no pose annotations and includes some borderline cases (e.g., tiny head fragments).

Since CIHP is a different domain than OCHuman, we report segmentation results on both.

\begin{table*}[tb]
    \vspace{-1em}
    \centering
    \newcolumntype{C}{>{\centering\arraybackslash}X}
    \caption{
    \textbf{Pose estimation -- comparison with state-of-the-art}.
    Best results in bold; second best underlined.
    $\ddagger$ results not reported.
    $\dagger$ uses a different detector.
    $\mathparagraph$ ignores \textit{small} instances in~COCO.\\
    Summary:
    PMPose improves MaskPose and sets the new SOTA for top-down approaches.
    BMPv2 improves BMPv1 \cite{BMPv1} and BMPv2+ significantly outperforms the previous SOTA.
    }
    \label{tab:SOTA-comparison-pose}
\begin{tabularx}{\linewidth}{@{} l || CC | CC | C }
        \toprule
        \multirow{2}{*}{Model} &  \multicolumn{2}{c|}{OCHuman} &  \multicolumn{2}{c|}{OCHuman-Pose}       & COCO  \\
                                & val AP & test AP & val AP & test AP       & val AP   \\

        \midrule
        \multicolumn{6}{l}{Methods without explicit detector (single stage + bottom-up)} \vspace{0.2em}\\
        
        DEKR       \cite{DEKR}  & 37.9 &  36.5 & ---$^{\ddagger}$ & ---$^{\ddagger}$        & 71.0    \\
        HQNet R-50$^{\mathparagraph}$ \cite{HQNet} & ---$^{\ddagger}$&  40.0 & ---$^{\ddagger}$& ---$^{\ddagger}$        & 69.5  \\
        CID-w48    \cite{CID} & 46.1 &  45.0 & ---$^{\ddagger}$ & ---$^{\ddagger}$    &  68.9  \\
        
        BUCTD \cite{BUCTD} & 48.3 &  47.4  & ---$^{\ddagger}$   & ---$^{\ddagger}$      & 74.8     \\

        \midrule
        \multicolumn{6}{l}{Top-down methods; bounding boxes from RTMDet-l \cite{RTMDet}} \vspace{0.2em}\\

        Sapiens 0.3b \cite{Sapiens}& 42.0 &  41.3  & 69.8 & 70.5        & 66.1    \\
        MIPNet$^{\dagger}$   \cite{MIPNet}& 42.0 &  42.5 & ---$^{\ddagger}$ & ---$^{\ddagger}$    & 76.3    \\
        
        ViTPose-B \cite{ViTPose} & 37.6 &  37.5 & 66.9 & 67.8     & 76.4 \\ 
        
        ViTPose*-B \cite{ViTPose} & 44.5 & 44.1 & 75.3 & 76.1     & 77.3 \\ 
        
        MaskPose-B \cite{BMPv1} & 46.6 & 46.6 &  77.1  & 78.0   & 76.8\\ 

        \rowcolor{LighterYellow}
        PMPose-B             & 47.9 & 48.2  & 78.9 & 80.0    & 76.9  \\ %
        
        \midrule
        \multicolumn{6}{l}{Iterative methods} \vspace{0.2em}\\

        BUCTD $2\times$ \cite{BUCTD}  & 48.8 &  48.3  & ---$^{\ddagger}$ & ---$^{\ddagger}$  & ---$^{\ddagger}$   \\
        
        BMPv1 $2\times$    \cite{BMPv1}          & 48.7 & 49.2 & 79.0 & 79.9     & 76.5    \\  
        
        \rowcolor{LighterYellow}
        BMPv2 $2\times$    & \underline{51.3} & \underline{51.5} &  \underline{82.4}   & \underline{83.4}       & \textbf{78.8}   \\ %
        \rowcolor{LighterYellow}
        BMPv2+ $2\times$    & \textbf{55.8} & \textbf{55.8} &  \textbf{85.8}   & \textbf{86.8}      & \underline{78.1}   \\        
        
        \bottomrule
    \end{tabularx}
    \vspace{-0.5em}
\end{table*}

\begin{table*}[tb]
    \centering
    \newcolumntype{C}{>{\centering\arraybackslash}X}
    \caption{
    \textbf{
    Detection and segmentation -- comparison with state-of-the-art}.
    Best results in bold; second best underlined.
    $\ddagger$ results not reported.
    BMPv2 improves BMPv1 \cite{BMPv1} especially in instance segmentation thanks to SAM-pose2seg.
    BMPv2+ results are the same as BMPv2, as the `plus' version differs only by an additional PMPose iteration.
    }
    \label{tab:SOTA-comparison-det}
    \begin{tabularx}{\linewidth}{@{} l|| CC | CC  | C  C | CC }
        \toprule
        \multirow{3}{*}{Model} &  \multicolumn{4}{c|}{OCHuman} &  \multicolumn{2}{c|}{OCHuman-Pose}  & \multicolumn{2}{c}{CIHP}    \\
                                & \multicolumn{2}{c}{val AP} & \multicolumn{2}{c|}{test AP} & val~AP & test~AP       & \multicolumn{2}{c}{val AP}   \\
                                & bbox & mask & bbox & mask & bbox & bbox & bbox & mask \\
    \midrule
    \multicolumn{9}{l}{Detectors} \vspace{0.2em}\\
    HRNet \cite{HRNet}           & 27.4 & 19.9 & 27.1 & 19.4 & 39.1 & 39.8 & 65.2 & 58.7 \\
    ConvNeXt \cite{ConvNeXt}     & 29.2 & 19.8 & 29.4 & 20.4 & 41.4 & 42.2 & 69.1 & 60.5 \\
    HQNet R-50 \cite{HQNet}      & 30.6 & 31.5 & 29.5 & 31.1 & ---$^{\ddagger}$ & ---$^{\ddagger}$ & ---$^{\ddagger}$ & ---$^{\ddagger}$ \\
    CoDETR SWIN-L \cite{coDETR}  & 29.0 & ---$^{\ddagger}$  & 29.6 & ---$^{\ddagger}$  & 41.6 & 42.5 & \textbf{73.5} & ---$^{\ddagger}$  \\
    RTMDet-l \cite{RTMDet}       & 31.1 & 27.1 & 30.0 & 26.2 & 45.7 & 45.7 & 69.5 & 63.9 \\ %
    
    \midrule
    \multicolumn{9}{l}{Pose-guided segmentation methods} \vspace{0.2em}\\
    Occlusion C\&P \cite{HumanPaste} & ---$^{\ddagger}$  & 28.9 & ---$^{\ddagger}$  & 28.3 & ---$^{\ddagger}$ & ---$^{\ddagger}$ & ---$^{\ddagger}$ & ---$^{\ddagger}$ \\
    ExPoSeg \cite{PoSeg}             & ---$^{\ddagger}$  & 26.4 & ---$^{\ddagger}$  & 26.8 & ---$^{\ddagger}$ & ---$^{\ddagger}$ & ---$^{\ddagger}$ & ---$^{\ddagger}$ \\
    Crowd-SAM \cite{CrowdSAM}        & ---$^{\ddagger}$  & 31.4 & ---$^{\ddagger}$  & ---$^{\ddagger}$ & ---$^{\ddagger}$ & ---$^{\ddagger}$ & ---$^{\ddagger}$ & ---$^{\ddagger}$ \\
    
    \midrule
    \multicolumn{9}{l}{Iterative methods} \vspace{0.2em}\\
    BMPv1 2$\times$ \cite{BMPv1} & \underline{36.3} & \underline{33.7} & \underline{35.9} & \underline{34.0} & \underline{49.0} & \underline{48.8} & \underline{69.7} & \underline{65.9} \\ %
    \rowcolor{LighterYellow}
    BMPv2 2$\times$              & \textbf{37.6} & \textbf{40.9} & \textbf{37.5} & \textbf{41.1} & \textbf{50.4} & \textbf{51.3} & 68.8 & \textbf{69.5} \\

    \bottomrule
    \end{tabularx}
    \vspace{-0.5em}
\end{table*}

\section{Experiments}
\label{sec:exp}

Unless stated otherwise, we use the following models.
RTMDet-L from~\cite{BMPv1} is used for detection.

PMPose-B is used for human pose estimation.
It is trained on a mixture of COCO~\cite{COCO}, MPII~\cite{MPII}, and AIC~\cite{AIC}.
For MPII and AIC, segmentation masks are generated using SAM2.1~\cite{SAM2} with ground-truth bounding boxes and poses as prompts.
Mask augmentation is critical for generalization and robustness to detected masks.

SAM-pose2seg is built on a SAM-hiera-B+ backbone with our adaptation (see \cref{sec:method-SAM}).
It is fine-tuned on COCO and CIHP~\cite{CIHP}.
Since CIHP training data lacks pose annotations, pseudo ground-truth poses are generated using an earlier PMPose-B variant together with ground-truth segmentation masks.

Training details (e.g., schedules and learning rates), together with training scripts for reproducibility, are provided in Suppl. on the \ProjectWebsite

\subsection{2D Pose Estimation}
\label{sec:exp-SOTA-pose}

\cref{tab:SOTA-comparison-pose} compares BMPv2 with the state-of-the-art in crowded human pose estimation.
We show a comparison on the OCHuman dataset for comparison with previous approaches.
More reliable comparison is on OCHuman-Pose with fixed annotations as explained in \cref{sec:data-OCH}.
Lastly, performance on COCO is shown for comparison in non-crowded scenarios.

Our new top-down method, PMPose-B, significantly outperforms the state-of-the-art among top-down methods in crowded scenes while keeping performance on COCO.
PMPose performs comparably to BUCTD~\cite{BUCTD}, which is an iterative method.
As shown in \cref{fig:PMPose-vs-ViTPose}, PMPose is superior to the compared pose models, setting a new state-of-the-art.

BMPv2 outperforms BMPv1~\cite{BMPv1} in all measures on all datasets. 
BMPv2+ is even better and is the first method to exceed 50 AP and 80 AP performance on the OCHuman test and the OCHuman-Pose test, respectively.

BMPv2+ has slightly worse results on COCO compared to BMPv2 for two reasons.
One, COCO contains many small instances (more than 30\% of instances have a bbox size smaller than 100px).
Mask refinement is not suitable for such small instances (most pose-guided segmentation methods such as~\cite{pose2seg,PoSeg} ignore small instances in evaluation), and looping SAM-pose2seg with PMPose amplifies the error.
Second, COCO contains annotation errors along the bounding box edges as shown in~\cite{ProbPose}.
Therefore, ProbPose-like trained PMPose performs poorly on such keypoints.

The last trend worth mentioning is the performance difference between OCHuman and OCHuman-Pose.
By annotating missing instances, the performance measure better corresponds to reality and shows the gap between detection error and pose estimation error.
BMPv2+ has almost the same performance as ViTPose with GT bboxes (85.8 vs. 86.4) on val set and even outperforms ViTPose with GT bboxes on the test set (86.8 vs. 86.2).
The results of ViTPose with GT bboxes are in \cref{tab:och-pose-gt-vs-det}.
These results show that, with the BMP loop, the gap between GT and detected boxes is much smaller than on the original OCHuman. This suggests that detection is no longer the dominant bottleneck, and that errors in OCHuman-Pose arise to a large extent from pose estimation rather than detection.

\subsection{Detection and Segmentation}
\label{sec:exp-SOTA-det}

Comparison with SOTA on detection and instance segmentation is in \cref{tab:SOTA-comparison-det}.
The first part of the table shows standard detectors, the middle part shows pose-guided instance segmentation models, and the bottom part compares against BMPv1.

BMPv2 outperforms BMPv1 in all measures. 
The most dramatic increase is in instance segmentation thanks to the improved SAM-pose2seg.%

The only measure where BMP is not SOTA is CIHP bounding-box AP.
Here BMPv2 even underperforms BMPv1.
The improved SAM-pose2seg is built for images with higher resolution and instances of people.
CIHP contains small instances similarly to COCO.
Further, CIHP contains many instances where only a very small part of the person is visible (e.g. the top of the head and no keypoints).
For masks on CIHP, BMPv2 is the best because the mask improvement for bigger instances outweighs wrongly detected small bounding boxes.

Since BMPv2+ converges after second run on PMPose (see \cref{sec:bmp-plus}), its results for detection and instance segmentation are the same as BMPv2.

\begin{table}[tb]
    \centering
    \newcolumntype{C}{>{\centering\arraybackslash}X}
    \caption{
    \textbf{Ablation of BMP components.}
    SAM-pose2seg improves mask AP, PMPose helps pose AP.
    SAM-pose2seg and PMPose in BMPv2+ bring dramatic pose improvements.
    BMPv1 with PMPose gives worse segmentation than standalone BMPv1, as the original prompting is optimized for MaskPose's confidence.%
    }
    \label{tab:abl-parts}
    \begin{tabularx}{\linewidth}{l|CC}
    \toprule
    \multirow{2}{*}{BMP version} &  \multicolumn{2}{c}{OCHuman} \\
                                 & pose AP & mask AP \\
    \midrule
     BMPv1                 & 49.2 & 34.0 \\
     BMPv1 + SAM-pose2seg  & 49.5 & \underline{40.0} \\
     BMPv1 + PMPose        & 50.2 & 30.8 \\
     BMPv2                 & \underline{51.5} & \textbf{41.1} \\
     BMPv2+                & \textbf{55.8} & \textbf{41.1} \\
    \bottomrule
    \end{tabularx}
    \vspace{-1.5em}
\end{table}

\subsection{Ablation Study}
\label{sec:exp-abl}

\cref{tab:abl-parts} shows contribution of SAM-pose2seg, PMPose and the `plus' version.
As expected, SAM-pose2seg improves instance segmentation even for BMPv1 due to its human-focused fine-tuning.
Similarly, PMPose improves pose estimation even in BMPv1.

The most dramatic improvement is BMPv2+ which loops SAM-pose2seg and PMPose.
Improvements in both models add more than 4 AP points on pose estimation.

Note that when SAM-pose2seg is used in BMPv1, it is prompted with BMPv1's complex prompt selection method for fair comparison.
Since MaskPose~\cite{BMPv1} does not predict visibility, using BMPv2 prompting method would bring unfairly bad results.

\subsubsection{PMPose Performance}
\label{sec:exp-abl-PMPose}

\begin{table}[tb]
\centering
\caption{
\textbf{PMPose vs. similar models using ViT backbone} \cite{ViT} but different decoders and training schemes.
Bounding boxes from RTMDet-L \cite{RTMDet}.
MaskPose$^\dagger$ was re-trained with an improved training schedule compared to the \cite{BMPv1}.
PMPose is the only model combining mask conditioning with probabilistic outputs, and it keeps SOTA performance both in crowds and in standard images.
}
\label{tab:pmpose-vs-rest}
\newcolumntype{C}{>{\centering\arraybackslash}X}
\begin{tabularx}{\linewidth}{@{}lcc|CC@{}}
    \toprule
    \multirow{2}{*}{Model} & mask & \multirow{2}{*}{$v+p_p$} & COCO & OCH \\
    & input & & val AP & test AP \\ 
    \midrule
    ViTPose-S \cite{ViTPose}   & \xmark & \xmark & 75.9 & 35.5 \\
    ViTPose-B \cite{ViTPose}   & \xmark & \xmark & 77.9 & 37.5 \\
    MaskPose-B \cite{BMPv1}    & \cmark & \xmark & 76.5 & 45.0 \\
    ProbPose-S \cite{ProbPose} & \xmark & \cmark & 76.6 & 37.7 \\
    \midrule
    MaskPose-S$^{\dagger}$     & \cmark & \xmark & 78.3 & 45.7 \\
    MaskPose-B$^{\dagger}$     & \cmark & \xmark & \textbf{79.8} & \underline{46.9} \\
    \rowcolor{LighterYellow}
    PMPose-S                   & \cmark & \cmark & 77.9 & 46.8 \\
    \rowcolor{LighterYellow}
    PMPose-B                   & \cmark & \cmark & \underline{79.4} & \textbf{48.2 }\\
    \bottomrule
    \end{tabularx}
\vspace{-1.5em}
\end{table}

The \cref{tab:pmpose-vs-rest} compares PMPose with ViTPose~\cite{ViTPose}, ProbPose~\cite{ProbPose}, and MaskPose~\cite{BMPv1}.

MaskPose introduced mask conditioning, which greatly improved pose estimation in crowded OCHuman scenes.
ProbPose introduced presence probability ($p_p$) and improved visibility ($v$) estimation. Its probabilistic keypoint representation -- presence probability, visibility, and calibrated probmaps -- provides richer information for downstream processing and SAM prompting.
As shown in \cref{tab:sam-pose2seg-prompt}, visibility estimation is especially useful for keypoint selection.

Mask conditioning, however, is a core component of the BMP loop.
It is necessary for reliable instance separation in crowded scenes, which makes ProbPose harder to use in the loop.
PMPose combines the strengths of both methods.
It predicts visibility, presence probability, and calibrated probmaps on mask-conditioned images.
These two components are complementary rather than conflicting.
As shown in \cref{tab:pmpose-vs-rest}, PMPose achieves SOTA performance on both crowded OCHuman images and standard COCO images.

The small drop from re-trained MaskPose-B to PMPose-B on COCO (79.8~$\rightarrow$~79.4) is caused by incorrect border-point annotations in COCO.
As explained in \cite{ProbPose}, COCO often places annotations inside the image even when the keypoint lies outside it.
See \cite{ProbPose}, especially the supplementary material, for details.

\subsubsection{Visibility Estimation for SAM-pose2seg}
\label{sec:exp-abl-vis}

\begin{table}[tb]
\centering
\caption{
\textbf{SAM prompting with keypoints selected by different metrics.}
For SAM~\cite{SAM2}, prompts are chosen using the prompting strategy from~\cite{BMPv1}, whereas SAM-pose2seg uses the top-3 points per metric.
These configurations correspond to the optimal settings for each model.
For confidence$^\dagger$, MaskPose~\cite{BMPv1} was used as PMPose does not return confidence.
Visibility is the best predictor of usefulness for keypoint prompting.
}
\label{tab:sam-pose2seg-prompt}
\newcolumntype{C}{>{\centering\arraybackslash}X}
\begin{tabularx}{\linewidth}{@{}l|CCC@{}}
    \toprule
    \multirow{2}{*}{Metric} & COCO & OCH & CIHP \\
     & val AP & test AP & val AP \\ 
    \midrule
    \multicolumn{4}{l}{SAM 2.1} \vspace{0.2em}\\
    conf.$^\dagger$ & 32.7 & 24.6 & 58.6 \\
    pres. prob. & \underline{38.2} & 23.0 & 61.8 \\
    exp. OKS & 35.4 & \underline{25.3} & \underline{64.5} \\
    \rowcolor{LighterYellow}
    vis & \textbf{41.2} &\textbf{29.5} & \textbf{71.6} \\
    \midrule
    \multicolumn{4}{l}{SAM-pose2seg} \vspace{0.2em}\\
    conf.$^\dagger$ & 38.3 & \underline{33.6} & 66.6 \\
    pres. prob. & 43.8 & 31.0 & 69.3 \\
    exp. OKS & \underline{43.9} & \textbf{34.7} & \underline{72.0} \\
    \rowcolor{LighterYellow}
    vis. & \textbf{44.6} & \textbf{34.7} & \textbf{72.7} \\
    \bottomrule
\end{tabularx}
\vspace{-1.5em}
\end{table}

\cref{tab:sam-pose2seg-prompt} shows how SAM 2.1~\cite{SAM2} and SAM-pose2seg behave when prompted with different keypoints.
PMPose predicts (1) presence probability, (2) expected OKS and (3) visibility for each keypoint.
All three variables, along with MaskPose's confidence~\cite{BMPv1} can be used to select keypoints for prompting.
In the experiment, SAM-pose2seg was fine-tuned for each variable for fair comparison.
As shown in \cref{tab:sam-pose2seg-prompt}, PMPose's visibility is the best predictor of keypoint usefulness.
PMPose is useful not only for improved localization but also because the information it provides about each keypoint can be used to select prompts more reliably.

Suppl. expands on this ablation and addresses the follow-up question of why visibility works the best.

\subsubsection{SAM-pose2seg qualitative examples}
\label{sec:exp-abl-vis-qual}

\begin{figure}[tb]
    \centering
    \begin{subfigure}{0.49\linewidth}
        \centering
        \includegraphics[width=\textwidth]{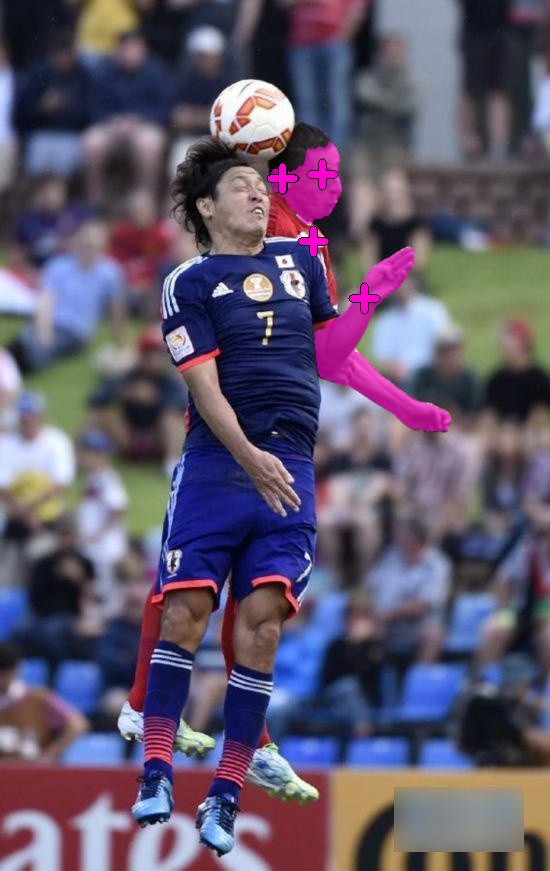}
    \end{subfigure}
    \hfill
    \begin{subfigure}{0.49\linewidth}
        \centering
        \includegraphics[width=\textwidth]{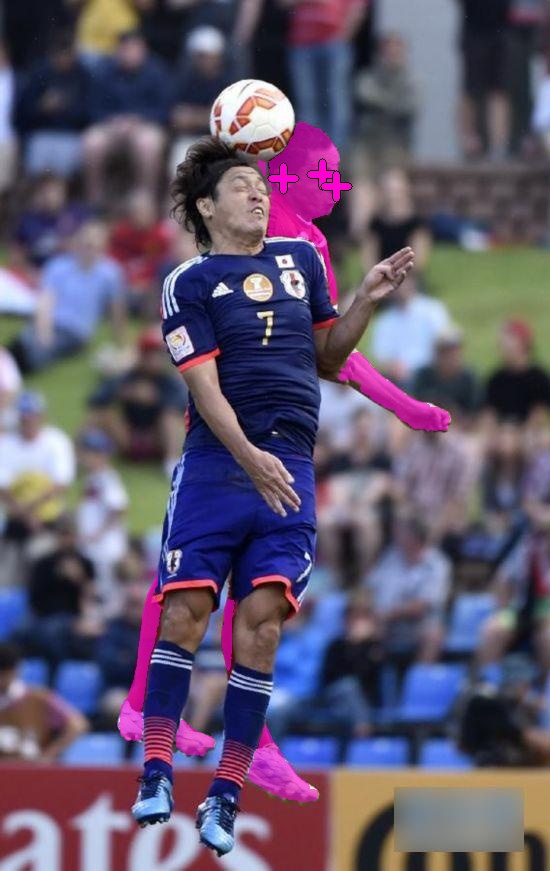}
    \end{subfigure}
    
    \begin{subfigure}{0.49\linewidth}
        \centering
        \includegraphics[width=\textwidth]{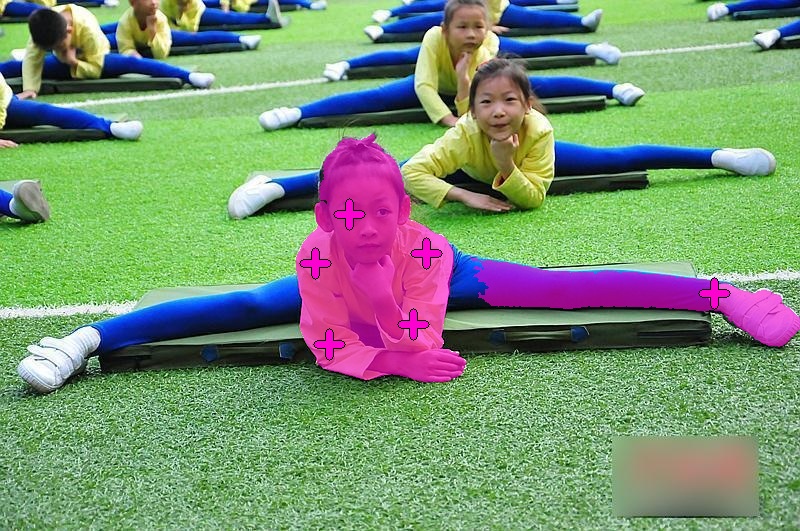}
    \end{subfigure}
    \hfill
    \begin{subfigure}{0.49\linewidth}
        \centering
        \includegraphics[width=\textwidth]{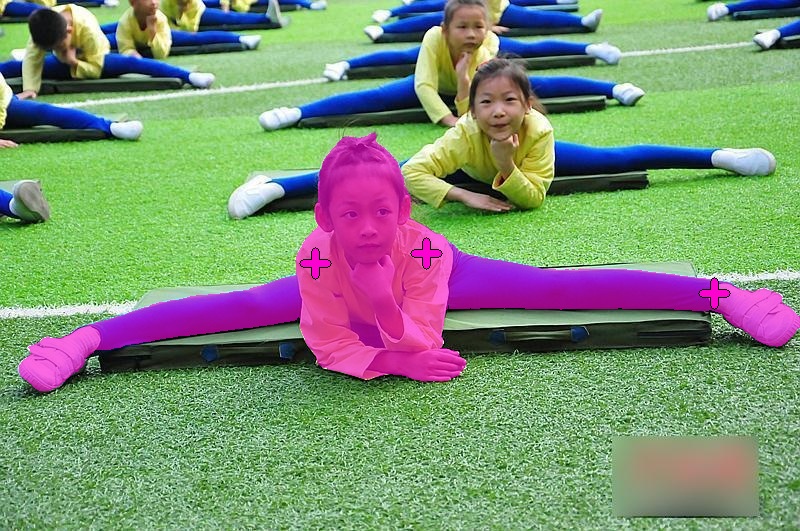}
    \end{subfigure}
    \caption{
    \textbf{Examples of reduced oversegmentation} on OCHuman.
    Left: mask refined with SAM 2.1 \cite{SAM2} using up to 6 keypoints (optimal strategy from \cite{BMPv1}).
    Right: mask refined with our SAM-pose2seg using up to 3 most visible keypoints.
    \vspace{-1.5em}
    }
    \label{fig:SAMp2s-oversegmentation}
\end{figure}

\cref{fig:SAMp2s-oversegmentation} illustrates a qualitative effect behind the mask AP gains of SAM-pose2seg in \cref{tab:SOTA-comparison-det,tab:abl-parts}.
When prompted in crowded scenes, the original SAM 2.1~\cite{SAM2} often oversegments adjacent people or merges them with nearby human-shaped regions.
After our human-focused fine-tuning described in \cref{sec:method-SAM}, the decoder becomes substantially better at separating nearby instances under pose guidance.
Because SAM-pose2seg predicts only the human class, it is less prone to the object ambiguity of the generic SAM decoder and therefore benefits from a much simpler prompting rule based on only the top-3 visible keypoints.

\subsubsection{Number of Iterations}
\label{sec:exp-abl-iter}

\pgfplotsset{
  bmpitersub/.style={
    width=1.0\linewidth,
    height=0.52\linewidth,
    xmin=0.5, xmax=4.5,
    xtick={1,2,3,4},
    xticklabels={$1\times$,$2\times$,$3\times$,$4\times$},
    xlabel={BMPv2 iterations},
    tick label style={font=\footnotesize},
    label style={font=\footnotesize},
    title style={font=\bfseries\small, yshift=-0.5ex, align=center},
    grid=major,
    major grid style={draw=black!8},
    axis line style={draw=black!65},
    clip=false,
  },
}

\colorlet{mapColor}{bmpColorBlue}
\colorlet{instColor}{bmpColorOrange}
\colorlet{instLabelColor}{bmpColorOrange!80!black}

\begin{figure}[t]
  \centering

\begin{tikzpicture}[baseline]
  \node (legcenter) at (2.2,0.08) {};

  \fill[instColor] ($(legcenter)+(-1.55,-0.07)$) rectangle +(0.16,0.16);
  \node[anchor=west, font=\footnotesize] at ($(legcenter)+(-1.40,0)$) {\# of instances};

  \draw[mapColor, line width=1.4pt] ($(legcenter)+(1.20,0)$) -- ($(legcenter)+(1.60,0)$);
  \fill[mapColor] ($(legcenter)+(1.40,0)$) circle (1.2pt);
  \node[anchor=west, font=\footnotesize] at ($(legcenter)+(1.65,0)$) {pose mAP};
\end{tikzpicture}

  \vspace{0.5em}

  \begin{tikzpicture}
    \begin{axis}[
      bmpitersub,
      ylabel={mAP},
      ymin=47.5, ymax=53.5,
      ytick={48,49,50,51,52,53},
    ]
      \addplot[
        ybar,
        bar width=14pt,
        draw=instColor,
        fill=instColor!75,
      ] coordinates {
        (1,49.0)
        (2,52.06)
        (3,52.16)
        (4,52.17)
      };

      \addplot[
        draw=mapColor,
        line width=1.4pt,
        mark=*,
        mark options={scale=0.65, fill=mapColor},
      ] coordinates {
        (1,48.4)
        (2,51.7)
        (3,51.8)
        (4,51.8)
      };

      \node[font=\scriptsize, text=instLabelColor, anchor=south] at (axis cs:1,49.05) {9912};
      \node[font=\scriptsize, text=instLabelColor, anchor=south] at (axis cs:2,52.11) {10532};
      \node[font=\scriptsize, text=instLabelColor, anchor=south] at (axis cs:3,52.21) {10551};
      \node[font=\scriptsize, text=instLabelColor, anchor=south] at (axis cs:4,52.22) {10554};
      
    \end{axis}
  \end{tikzpicture}

  \caption{
    \textbf{Impact of BMP iterations on the number of predictions and pose estimation performance} on OCHuman val \cite{pose2seg}.
    After second iteration, additional iterations bring only marginal instance gains and just a few new instances.
  }
  \label{fig:bmp-iterations-barline}
\end{figure}

\cref{fig:bmp-iterations-barline} shows that the main benefit of looping appears already in the second BMP iteration.
Pose mAP increases from 48.4 to 51.7, while the number of predictions rises from 9912 to 10532.
These additional detections are typically previously ignored instances which become recoverable once earlier predictions are blacked out.
Further iterations bring only diminishing returns: the third pass adds only 19 predictions and yields no meaningful mAP improvement, while the fourth adds only 3 predictions on the whole OCHuman validation set.
By that point, nearly all foreground pixels have already been blacked-out and the detector does not discover any new instances.

\subsubsection{Runtime Analysis}
\label{sec:exp-abl-runtime}

\cref{tab:runtime-per-parts} breaks down the inference cost of individual BMP components.
The most expensive part is the SAM backbone, followed by the detector, but both are executed only once per image.
In contrast, PMPose and the SAM decoder are evaluated per instance, yet each call is relatively cheap, which makes iterative refinement practical.
This explains the near-linear scaling in \cref{fig:bmp-runtime-scaling-measured}: runtime grows with the number of people, but the slope is controlled mainly by the lightweight per-instance stages rather than by repeatedly recomputing the heavy image-level features.

The broader comparison in \cref{fig:bmp-runtime-scaling-measured} and Suppl. also clarifies why we do not place large foundation models inside the loop.
Even Sapiens 0.3b~\cite{Sapiens} is slower than two BMP iterations, while full SAM-3D-Body~\cite{SAM-3D-Body} is substantially more expensive than the entire BMP pipeline.
For that reason, SAM-3D-Body is used only once as a downstream 3D prediction stage rather than as a recurrent component of the refinement loop.

\begin{table}[tb]
\centering
\setlength{\tabcolsep}{3.5pt} %
\caption{
\textbf{Runtime of BMP components}
measured on a single P100 with 16 GB memory and batch size 1, time in ms.
The second column shows if the component is per-image or per-instance.
The most expensive components run only once per image.
}
\label{tab:runtime-per-parts}
\begin{tabular}{@{}lc|rrrr@{}}
    \toprule
    Model & img & \multicolumn{1}{r}{p05} & \multicolumn{1}{r}{p50} & \multicolumn{1}{r}{avg} & \multicolumn{1}{r}{p95} \\
    \midrule
    RTMDet-L \cite{RTMDet} & \cmark & 43.3 & 47.8 & 49.4 & 60.3 \\
    PMPose-B & \xmark & 27.5 & 32.9 & 34.0 & 44.6 \\ 
    SAM bck. & \cmark & 139.8 & 140.1 & 140.1 & 140.6 \\
    SAM dec. & \xmark & 6.8 & 7.0 & 7.7 & 10.3 \\
    \midrule
    SAM-3DB \cite{SAM-3D-Body} & \xmark & 1347.3 & 1486.5 & 1471.2 & 1561.7 \\
    Sapiens 0.3b \cite{Sapiens} & \xmark & 548.3 & 562.0 & 565.4 & 583.1 \\ 
    \bottomrule
\end{tabular}
\vspace{-1.5em}
\end{table}

\pgfplotsset{
  bmpruntimesub/.style={
    width=1.0\linewidth,
    height=0.7\linewidth,
    xmin=1, xmax=7,
    xtick={1,2,3,4,5,6,7,8,9,10},
    xlabel={\# processed instances},
    ylabel={Runtime [ms]},
    tick label style={font=\footnotesize},
    label style={font=\footnotesize},
    title style={font=\bfseries\small, yshift=-0.5ex, align=center},
    legend style={
      font=\footnotesize,
      draw=none,
      fill=none,
      at={(0.5,1.03)},
      anchor=south,
      legend columns=2,
      /tikz/every even column/.append style={column sep=0.7em},
    },
    grid=major,
    major grid style={draw=black!8},
    axis line style={draw=black!65},
    clip=false,
  },
}

\colorlet{bmpRuntimeA}{bmpColorBlue}
\colorlet{bmpRuntimeB}{bmpColorOrange}
\colorlet{bmpRuntimeC}{bmpColorGreen}
\colorlet{bmpRuntimeD}{bmpColorMagenta}
\colorlet{bmpRuntimeS}{black!50}

\begin{figure}[t]
  \centering
  \begin{tikzpicture}
    \begin{axis}[
      bmpruntimesub,
      ymin=180, ymax=1080,
      ytick={200,400,...,1000},
    ]

      \addplot+[
        draw=bmpRuntimeA!60!white,
        line width=0.8pt,
        dashed,
        mark=null,
      ] table[
        x=n, y=med, row sep=\\,
      ]{
      n med \\
      1 226.50 \\
      2 268.96 \\
      3 311.42 \\
      4 353.88 \\
      5 396.34 \\
      6 438.80 \\
      7 481.26 \\
      };
      \addlegendentry{BMPv2\hphantom{+} 1$\times$ (theory)}

      \addplot[
        name path=bmpone_low,
        draw=none,
        forget plot,
      ] table[
        x=n, y=low, row sep=\\,
      ]{
      n low \\
      1 260.94 \\
      2 295.80 \\
      3 336.67 \\
      4 374.97 \\
      5 429.67 \\
      6 458.23 \\
      7 491.15 \\
      };

      \addplot[
        name path=bmpone_high,
        draw=none,
        forget plot,
      ] table[
        x=n, y=high, row sep=\\,
      ]{
      n high \\
      1 340.48 \\
      2 404.36 \\
      3 472.70 \\
      4 571.31 \\
      5 633.52 \\
      6 755.89 \\
      7 757.44 \\
      };

      \addplot[
        fill=bmpRuntimeA,
        fill opacity=0.2,
        draw=none,
        forget plot,
      ] fill between[of=bmpone_high and bmpone_low];

      \addplot+[
        draw=bmpRuntimeA,
        line width=1.2pt,
        mark=*,
        mark options={scale=0.7, fill=bmpRuntimeA},
      ] table[
        x=n, y=med, row sep=\\,
      ]{
      n med \\
      1 273.40 \\
      2 323.32 \\
      3 377.54 \\
      4 435.98 \\
      5 496.77 \\
      6 551.96 \\
      7 617.478 \\
      };
      \addlegendentry{BMPv2\hphantom{+} 1$\times$}

      \addplot+[
        draw=bmpRuntimeB,
        line width=0.8pt,
        mark=null,
        dashed,
      ] table[
        x=n, y=med, row sep=\\,
      ]{
      n med \\
      1 261.94 \\
      2 339.84 \\
      3 417.74 \\
      4 495.64 \\
      5 573.54 \\
      6 651.44 \\
      7 729.34 \\
      };
      \addlegendentry{BMPv2+ 1$\times$ (theory)}

      \addplot[
        name path=bmpone_low,
        draw=none,
        forget plot,
      ] table[
        x=n, y=low, row sep=\\,
      ]{
      n low \\
      1 317.88 \\
      2 385.85 \\
      3 436.98 \\
      4 510.48 \\
      5 589.44 \\
      6 640.67 \\
      7 720.89 \\
      };

      \addplot[
        name path=bmpone_high,
        draw=none,
        forget plot,
      ] table[
        x=n, y=high, row sep=\\,
      ]{
      n high \\
      1 384.44 \\
      2 493.58 \\
      3 628.39 \\
      4 745.94 \\
      5 863.18 \\
      6 993.09 \\
      7 1070.06 \\
      };

      \addplot[
        fill=bmpRuntimeB,
        fill opacity=0.15,
        draw=none,
        forget plot,
      ] fill between[of=bmpone_high and bmpone_low];

      \addplot+[
        draw=bmpRuntimeB,
        line width=1.2pt,
        mark=*,
        mark options={scale=0.7, fill=bmpRuntimeB},
      ] table[
        x=n, y=med, row sep=\\,
      ]{
      n med \\
      1 342.89 \\
      2 426.04 \\
      3 513.02 \\
      4 600.98 \\
      5 699.61 \\
      6 785.55 \\
      7 869.20 \\
      };
      \addlegendentry{BMPv2+ 1$\times$}

    \end{axis}
  \end{tikzpicture}

  \caption{
    \textbf{BMP runtime scaling with the number of processed instances.}
    Solid lines: median, shaded regions: the p05 to p95 range.
    Dashed lines: the theoretical runtime as a sum of components from \cref{tab:runtime-per-parts}. 
    Measured on a P100 GPU with 16GB memory and batch size 1 on OCHuman.
    Images with $>$7 detected instances are too rare for statistical evaluation.
    Comparison with other models is in the Suppl.
  }
  \label{fig:bmp-runtime-scaling-measured}
\end{figure}

\subsection{3D Pose Estimation}
\label{sec:exp-3D}

\cref{tab:sam3d-body-comparison} shows that BMP is suitable for SAM-3D-Body~\cite{SAM-3D-Body} prompting.
SAM-3D-Body needs a prompt for mesh reconstruction.
The prompt could be a bounding box, a mask, or a 2D pose.
Our experiments showed that pose does not help nearly as much as segmentation masks.
When SAM-3D-Body is prompted with BMPv2 masks, predicted 3D meshes are of comparable quality to 2D poses.
This suggests that precise masks simplify both 2D and 3D pose estimation in crowds.

\begin{figure*}[tb]
    \centering
    \begin{subfigure}{0.329\linewidth}
        \centering
        \includegraphics[width=\textwidth]{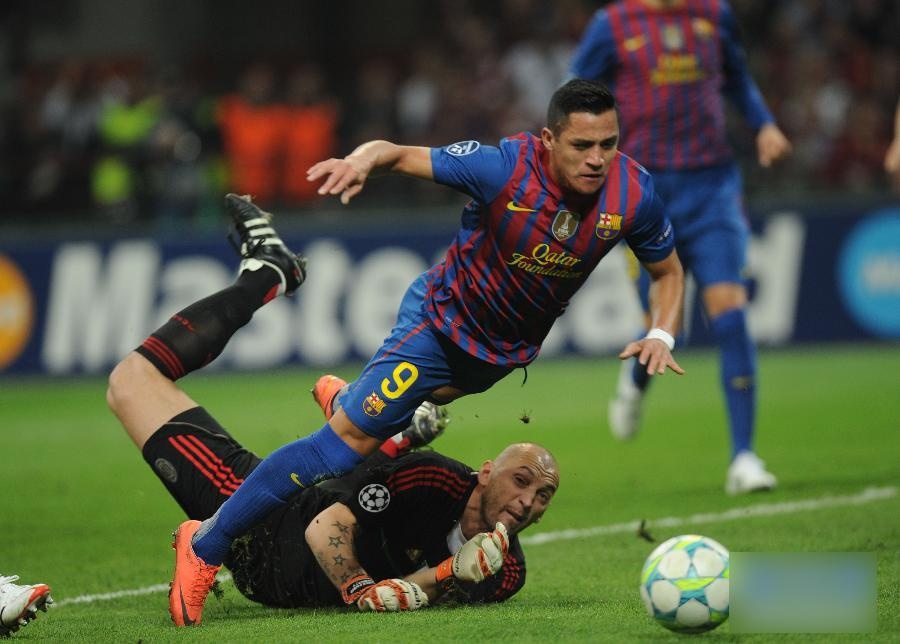}
    \end{subfigure}
    \hfill
    \begin{subfigure}{0.329\linewidth}
        \centering
        \includegraphics[width=\textwidth]{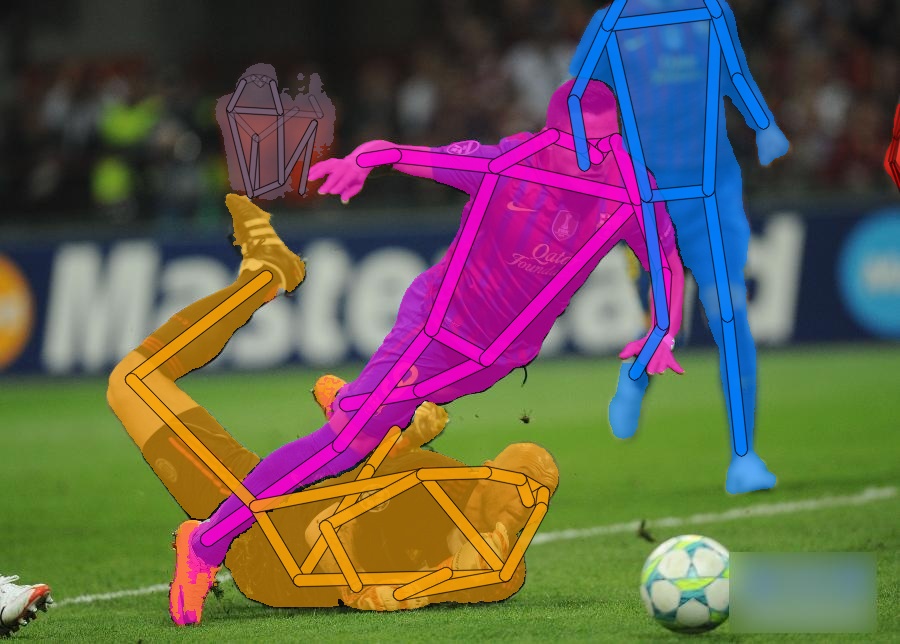}
    \end{subfigure}
    \hfill
    \begin{subfigure}{0.329\linewidth}
        \centering
        \includegraphics[width=\textwidth]{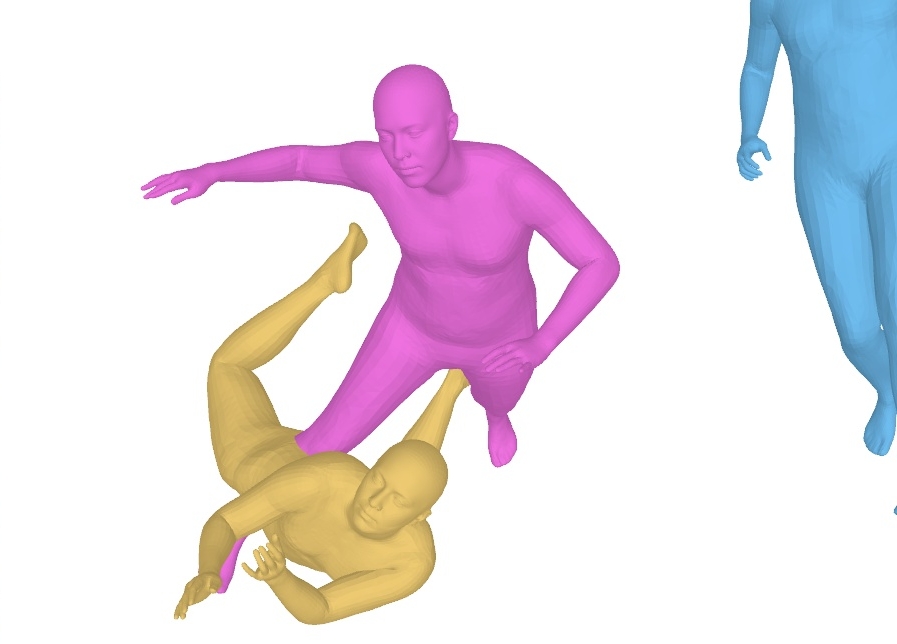}
    \end{subfigure}

    \begin{subfigure}{0.329\linewidth}
        \centering
        \includegraphics[width=\textwidth,height=6.0cm]{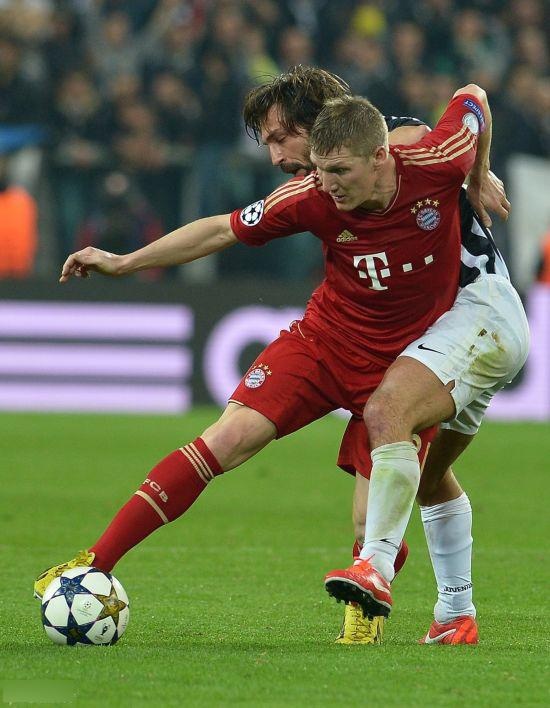}
    \end{subfigure}
    \hfill
    \begin{subfigure}{0.329\linewidth}
        \centering
        \includegraphics[width=\textwidth,height=6.0cm]{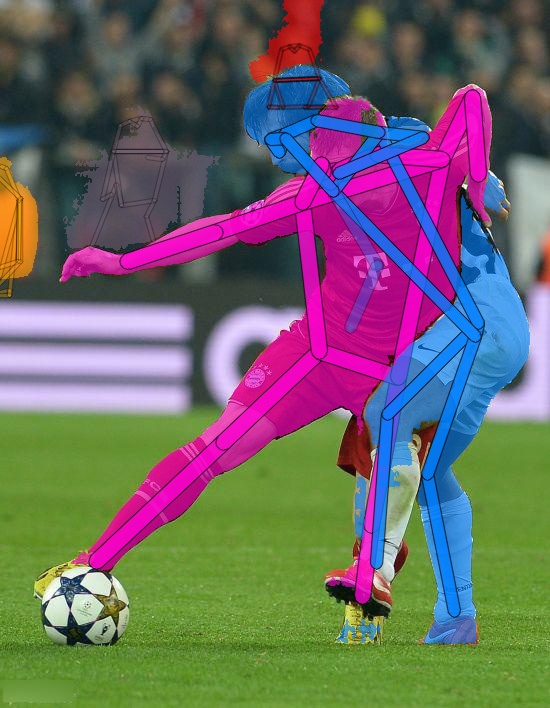}
    \end{subfigure}
    \hfill
    \begin{subfigure}{0.329\linewidth}
        \centering
        \includegraphics[width=\textwidth,height=6.0cm]{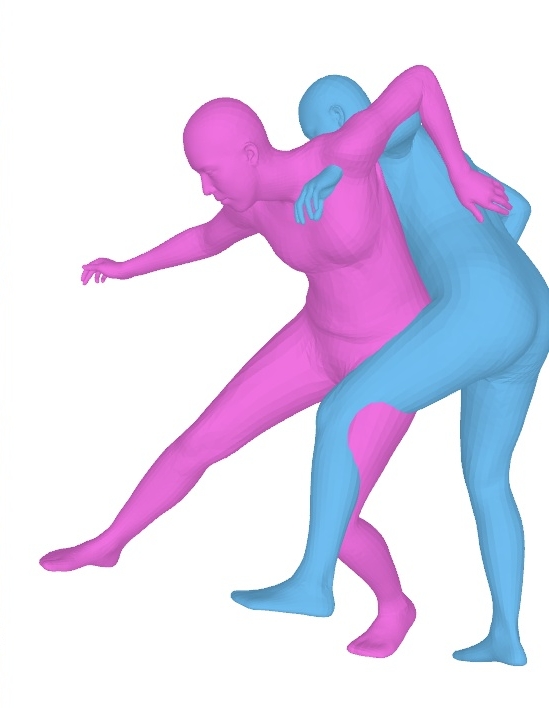}
    \end{subfigure}

    \begin{subfigure}{0.329\linewidth}
        \centering
        \includegraphics[width=\textwidth]{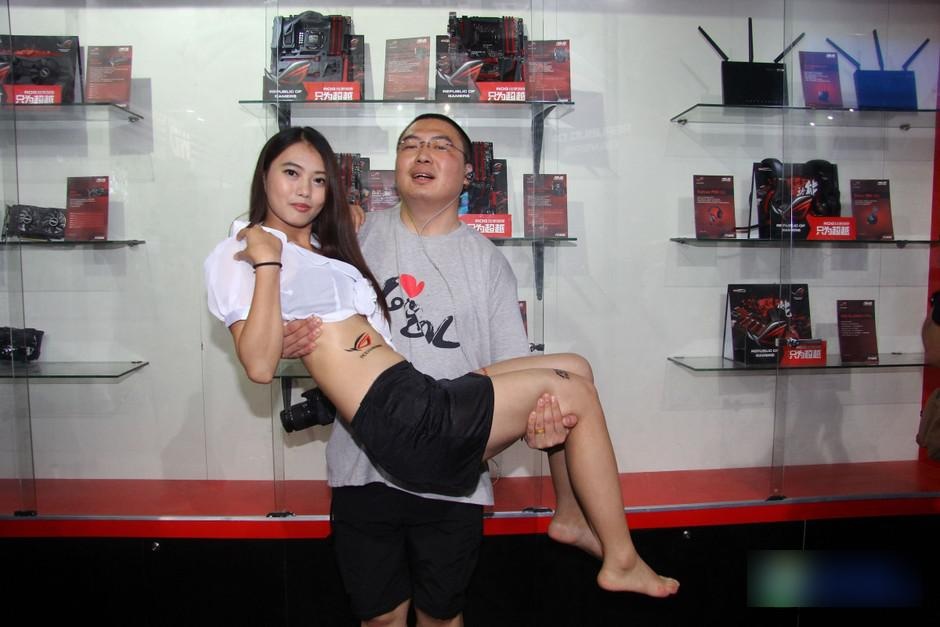}
    \end{subfigure}
    \hfill
    \begin{subfigure}{0.329\linewidth}
        \centering
        \includegraphics[width=\textwidth]{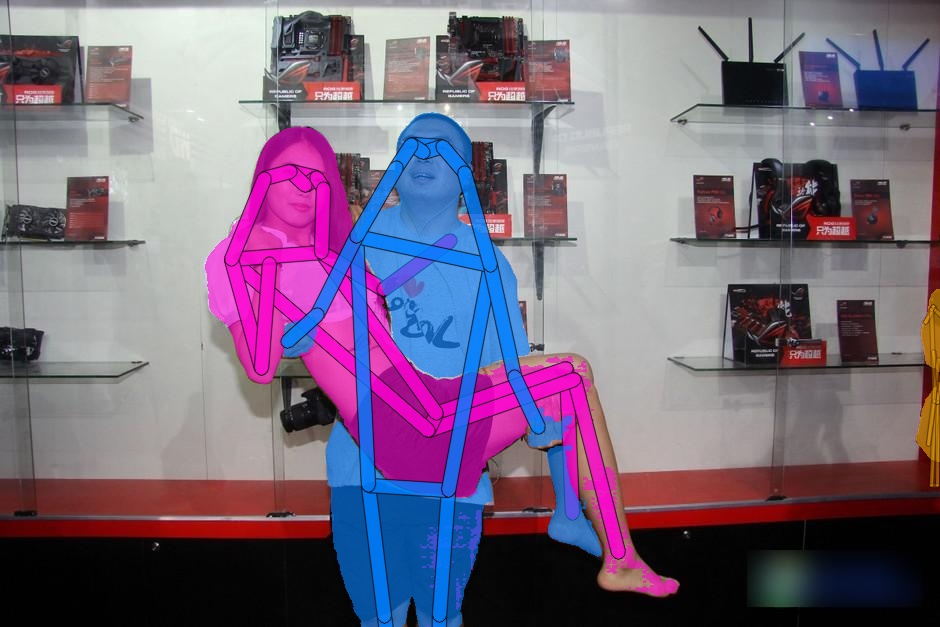}
    \end{subfigure}
    \hfill
    \begin{subfigure}{0.329\linewidth}
        \centering
        \includegraphics[width=\textwidth]{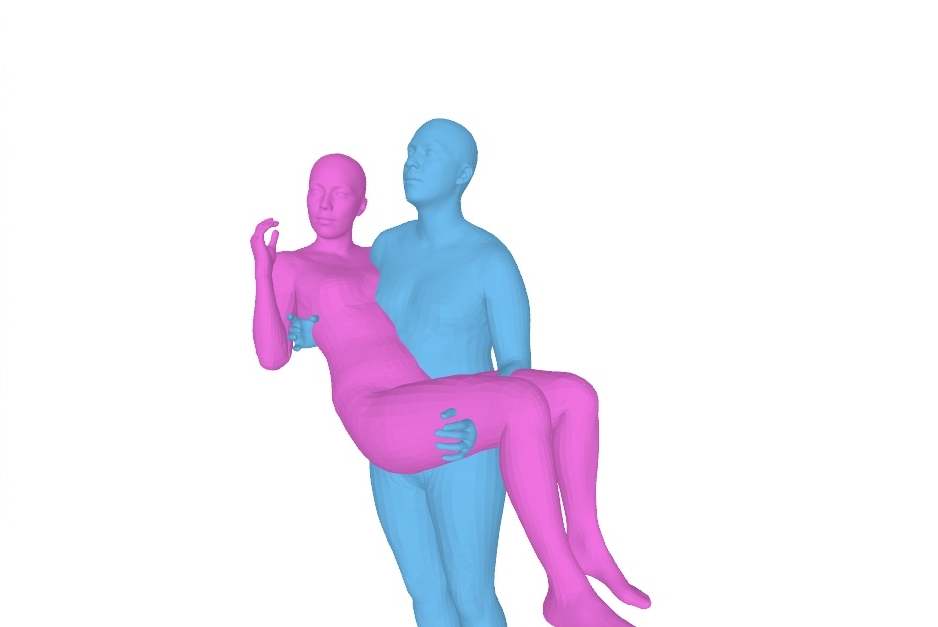}
    \end{subfigure}
    \caption{
    \textbf{Additional qualitative results} from OCHuman \cite{pose2seg}.
    }
    \label{fig:good-examples}
\end{figure*}

Note that \cref{tab:sam3d-body-comparison} measures the reprojection of 3D keypoints back to the 2D image, as there is no 3D ground truth for extremely crowded scenarios such as OCHuman.
The reprojection measure is good enough to measure the crude correctness of the pose but cannot measure the depth ambiguity.
However, depth ambiguity is usually not as large a problem in multi-body interactions as the correct assignment of limbs to individuals.
Qualitative examples are in \cref{fig:teaser,fig:good-examples}.

\begin{table}[t]
    \centering
    \newcolumntype{C}{>{\centering\arraybackslash}X}

    \caption{
    \textbf{Prompting methods for SAM-3D-Body}~\cite{SAM-3D-Body} on OCHuman test.
    Pose AP is the reprojection score of the 3D pose.
    Prompting with BMPv2 gives significantly better 3D poses thanks to much better instance masks. 
    }
    \label{tab:sam3d-body-comparison}
\begin{tabularx}{\linewidth}{@{}lc|C C@{}}
        \toprule
        \multirow{2}{*}{Detector} & \multirow{2}{*}{masks}  & OCHuman & OCH-Pose  \\
                                 &                        & 3D pose AP & 3D pose AP  \\
        \midrule
        ViTDet-H~\cite{ViTDet} & \xmark & 36.8  & 59.6 \\ 
        RTMDet-L~\cite{RTMDet} & \xmark & 38.5  & 68.0  \\ 
        BMP ~\cite{BMPv1} & \xmark & \underline{39.9}  & \underline{68.3}   \\ 
        \rowcolor{LighterYellow}
        BMPv2                      & \xmark & \textbf{46.4} & \textbf{75.8}  \\ 
        \midrule 
        RTMDet-L~\cite{RTMDet}       & \cmark & 40.0 & 69.5   \\
        BMP ~\cite{BMPv1}   & \cmark & \underline{45.6}  & \underline{74.9}  \\
        \rowcolor{LighterYellow}
        BMPv2               & \cmark & \textbf{54.8}  & \textbf{85.2}  \\
        \bottomrule
    \end{tabularx}
    \vspace{-1.5em}
\end{table}

\subsection{Robustness to Domain Shift}
\label{sec:exp-infants}

We evaluate BMP generalization on an unseen private dataset from~\cite{infants}.
The most challenging subset involves close adult-infant interactions, where standard pose estimators struggle to separate individuals.
The dataset is further challenging because (1) adults are often only partially visible, which degrades performance of top-down methods, and (2) infants are absent from standard training sets, causing a strong domain shift.
Due to privacy constraints, infant datasets are rarely shared, and models are typically not trained on infants.
As the dataset is private, none of the evaluated models -- including SAM -- were trained on its images, resulting in substantial domain shift.

\begin{figure*}[tb]
    
    \begin{subfigure}{0.329\linewidth}
        \centering
        \includegraphics[width=\textwidth]{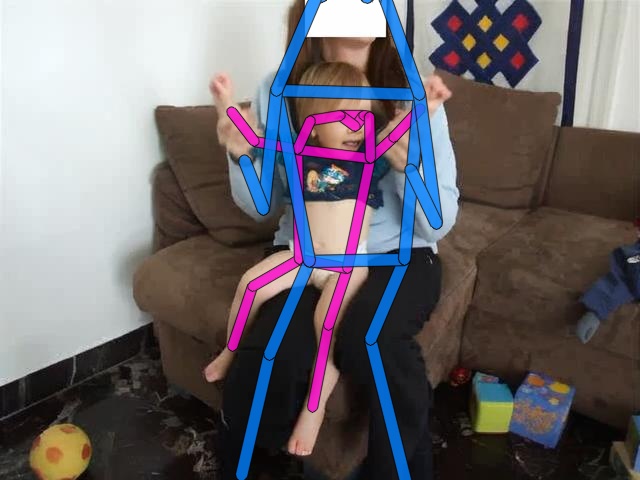}
    \end{subfigure}
    \hfill
    \begin{subfigure}{0.329\linewidth}
        \centering
        \includegraphics[width=\textwidth]{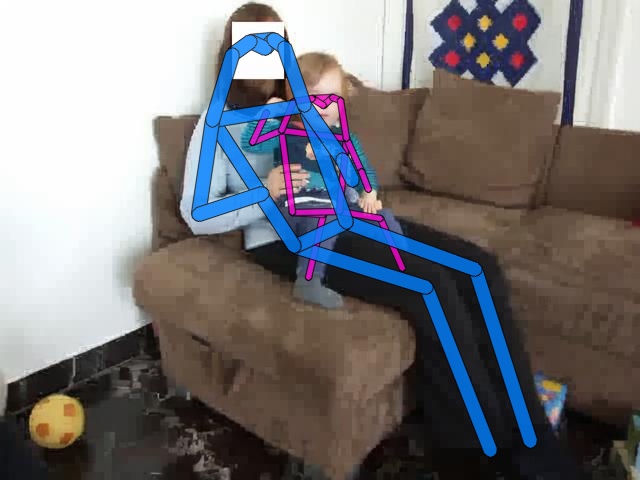}
    \end{subfigure}
    \hfill
    \begin{subfigure}{0.329\linewidth}
        \centering
        \includegraphics[width=\textwidth]{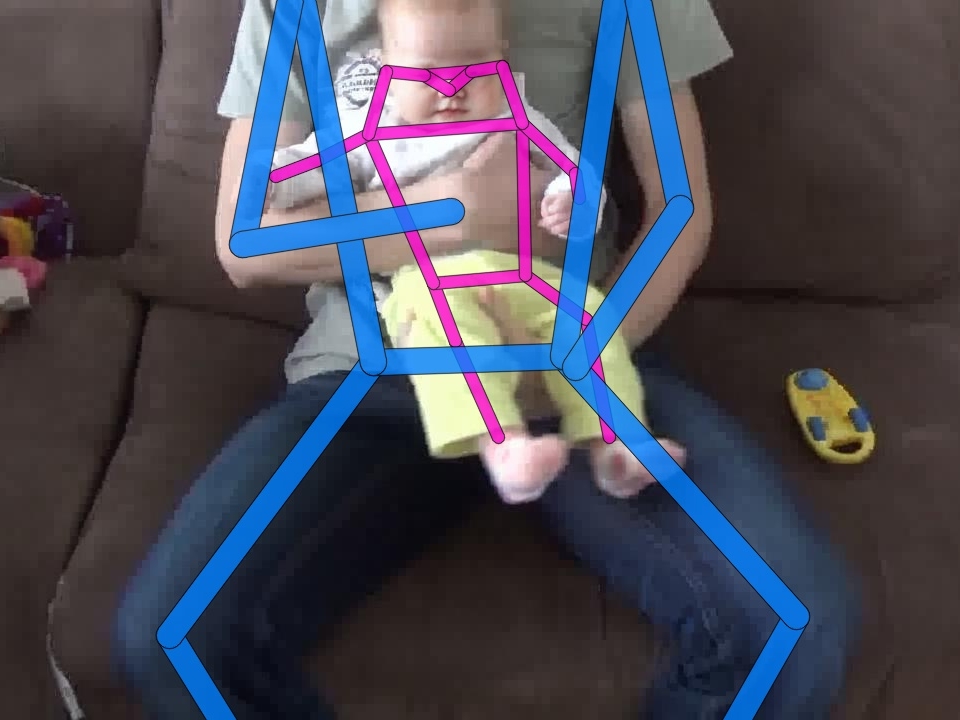}
    \end{subfigure}
    
    \caption{
    \textbf{BMPv2 results on a non-public dataset} from~\cite{infants}.
    BMPv2 is able to correctly disentangle individuals even in an unseen domain.
    Thanks to PMPose, keypoints outside of the image are not predicted inside.
    Results are shown only for authorized scenes; the dataset itself is more diverse than these examples.
    Infants anonymized with BLANKET~\cite{blanket}.
    }
    \vspace{-1em}
    \label{fig:infants-examples}
\end{figure*}

\cref{tab:infants,fig:infants-examples} report quantitative and qualitative results on the challenging lap subset.
Both BMPv1 and BMPv2 significantly outperform off-the-shelf SOTA (RTMDet~\cite{RTMDet} + ViTPose~\cite{ViTPose}).
PMPose, through probabilistic modeling, correctly predicts keypoints outside the image and avoids mixing individuals.
The dominant failure mode is missed detections in some frames rather than incorrect association.
Since evaluation is frame-based, such failures could be mitigated with temporal consistency in video.

BMPv2 achieves strong performance in complex multi-body scenes and is suitable when accuracy is the primary goal.
For time-constrained applications, PMPose, especially PMPose-s, offers a favorable trade-off between accuracy and speed.

\begin{table}[tb]
    \centering
    \newcolumntype{C}{>{\centering\arraybackslash}X}
    \caption{
    \textbf{Robustness to domain shift} on a non-public application dataset from~\cite{infants}.
    PMPose performs similarly to ViTPose*~\cite{ViTPose} while providing more information about each keypoint.
    Both BMPv1~\cite{BMPv1} and BMPv2 significantly improve over top-down methods.
    The private dataset guarantees that the data are from an unseen domain.
     }
    \label{tab:infants}
    \begin{tabularx}{\linewidth}{X  r}
    \toprule
    Method & pose AP \\
    \midrule
    RTMDet-l~\cite{RTMDet} + ViTPose-B~\cite{ViTPose}         & 73.3 \\
    RTMDet-l~\cite{RTMDet} + ViTPose*-B~\cite{ViTPose}        & 76.9 \\
    \rowcolor{LighterYellow}
    RTMDet-l~\cite{RTMDet} + PMPose-B                         & 76.4 \\
    BMPv1 2$\times$~\cite{BMPv1}                              & \underline{83.2} \\
    \rowcolor{LighterYellow}
    BMPv2 2$\times$                                           & \textbf{83.3} \\
    \bottomrule
    \end{tabularx}
    \vspace{-1.5em}
\end{table}

\section{Conclusions}
\label{sec:concl}

We present BBoxMaskPose v2 (BMPv2), an improved version of the BBoxMaskPose loop~\cite{BMPv1}.
Performance gains come from a stronger 2D pose estimator, PMPose, and a pose-guided adaptation of SAM for human instance segmentation, SAM-pose2seg.
BMPv2 outperforms BMPv1 across all metrics.
It is the first method to exceed 50 AP on OCHuman and also maintains state-of-the-art performance on COCO.

Key findings are:
\begin{enumerate}
    \item Unifying MaskPose~\cite{BMPv1} and ProbPose~\cite{ProbPose} yields PMPose, a more accurate and expressive model improving performance in both standard and crowded scenes.

    \item Adapting SAM-pose2seg~\cite{SAM-pose2seg-report} for pose-guided human instance segmentation significantly improves the BMP loop, beyond mask precision alone.

    \item PMPose and SAM-pose2seg enable an additional pose–mask refinement loop, forming BMPv2+. Pose estimation improves substantially, as the second PMPose iteration operates on masks refined by SAM-pose2seg. BMPv2+ sets a new SOTA on OCHuman.

    \item Poor performance on OCHuman~\cite{pose2seg} is caused by missing annotations. The proposed OCHuman-Pose dataset adds previously ignored instances to better reflect real scenes. On OCHuman-Pose, models using ground-truth and detected boxes have similar performance, indicating that pose estimation, not detection, is the main bottleneck.

    \item Robust 2D prompts are key for accurate 3D human pose estimation in crowded scenes. With BMPv2, state-of-the-art prompted 3D models remain effective on OCHuman even under severe instance overlap.

\end{enumerate}

\clearpage

\newpage
\balance

\subsubsection*{Funding}

This work was supported by
the Ministry of the Interior of the Czech Republic under project No. VL01010012 (FACIS2) within the IMPAKT 2 programme, %
the European Union’s Digital Europe Programme under Contract No. 101158609, %
BMK, BMAW, and the State of Upper Austria within the framework of the SCCH competence center INTEGRATE [Project 1.6 TFI (Transferable Intelligence)], which is part of the FFG COMET Competence Centers for Excellent Technologies Programme, %
and the Czech Technical University student grant SGS26/074/OHK3/1T/13.

\bibliography{sn-bibliography}%

\end{document}